\documentclass[letterpaper]{article} 
\usepackage[preprint]{aaai2027} 
\usepackage[hyphens]{url} 
\usepackage{graphicx} 
\usepackage{natbib} 
\usepackage{caption} 
\usepackage{amsmath}
\usepackage{amssymb}
\usepackage{booktabs}
\usepackage{multirow}
\usepackage{algorithm}
\usepackage{algorithmic}
\graphicspath{{figures/}}

\newcommand{\dMLLM}{dMLLM}
\newcommand{\confimg}{c^{I}_{t,i}}
\newcommand{\pimg}{p^{I}_{t,i}}
\newcommand{\pabl}{p^{A}_{t,i}}
\newcommand{\gvis}{g_{t,i}}
\newcommand{\Himg}{H^{\mathrm{img}}}

\title{Look Before You Commit: Visual-Evidence-Guided Decoding for
       Multimodal Diffusion Language Models}

\author{
  Tongxiao Hu, Y. Zhao, Q. Zhao, X. Yuan, X. Chen, P. Cheng, X. Zhang
}
\affiliations{}

\begin{document}

\maketitle

\begin{abstract}
Diffusion multimodal large language models rely on confidence to decide when
to commit candidate tokens during generation. In multimodal tasks, however,
confidence cannot distinguish whether a candidate originates from image
evidence or the language prior. We perform paired forward ablation and find
that nearly nine out of ten high-confidence commits are insensitive to the
image, while degenerate and hallucination samples exhibit significant
increases in the entropy of deep-layer attention organization. Based on this
diagnosis, we propose CD-APC, a training-free
decoding method. CD-APC performs evidence-adjusted scoring within a
base-supported pool and uses dual gating based on model acceptability and
contrastive concentration to decide whether to commit, together with an
inverse-window mechanism that adapts to inter-step difficulty. Across two
backbones and five benchmarks, CD-APC consistently improves performance,
indicating that stable visual evidence improves model outputs.
\end{abstract}




\section{Introduction}
\label{sec:intro}
Diffusion multimodal large language models (\dMLLM{}s) extend the masked
diffusion framework to image-conditioned text generation, and their parallel
decoding capability has demonstrated strong performance in visual question
answering, multimodal reasoning, and vision--language generation
\citep{wu2025fastdllmv2,jiang2025d2cache,kang2025parallelbench,
wei2025slowfast,wang2025d2f}. Unlike autoregressive models that decode one token
at a time, a \dMLLM{} predicts candidates for all pending positions
simultaneously at each denoising step. The decoder must therefore select a
subset $\mathcal{S}_t$ from the mask set $\mathcal{M}_t$ to write into the
sequence, referred to as \emph{commit scheduling}. Since committed tokens
immediately become part of the context for subsequent denoising steps, an early
erroneous commit propagates through later predictions and cannot be corrected
once written. Commit scheduling therefore plays a central role in preserving
visual evidence throughout the denoising process.

Existing \dMLLM{}s perform commit scheduling primarily according to candidate
confidence \citep{kang2025parallelbench}. Furthermore, recent studies have refined this confidence-based
paradigm along three directions: (i) adaptive-budget methods dynamically
allocate the commit budget according to distributional entropy or semantic
granularity \citep{wu2025fastdllm,lu2026adablock}; (ii) deferred-commitment
methods postpone low-confidence positions until additional context becomes
available \citep{shu2026dcd}; and (iii) position--step-prior methods use
denoising progress and position as scheduling priors \citep{kim2026thinking}.

\begin{figure}[t]
  \centering
  \includegraphics[width=0.85\columnwidth]{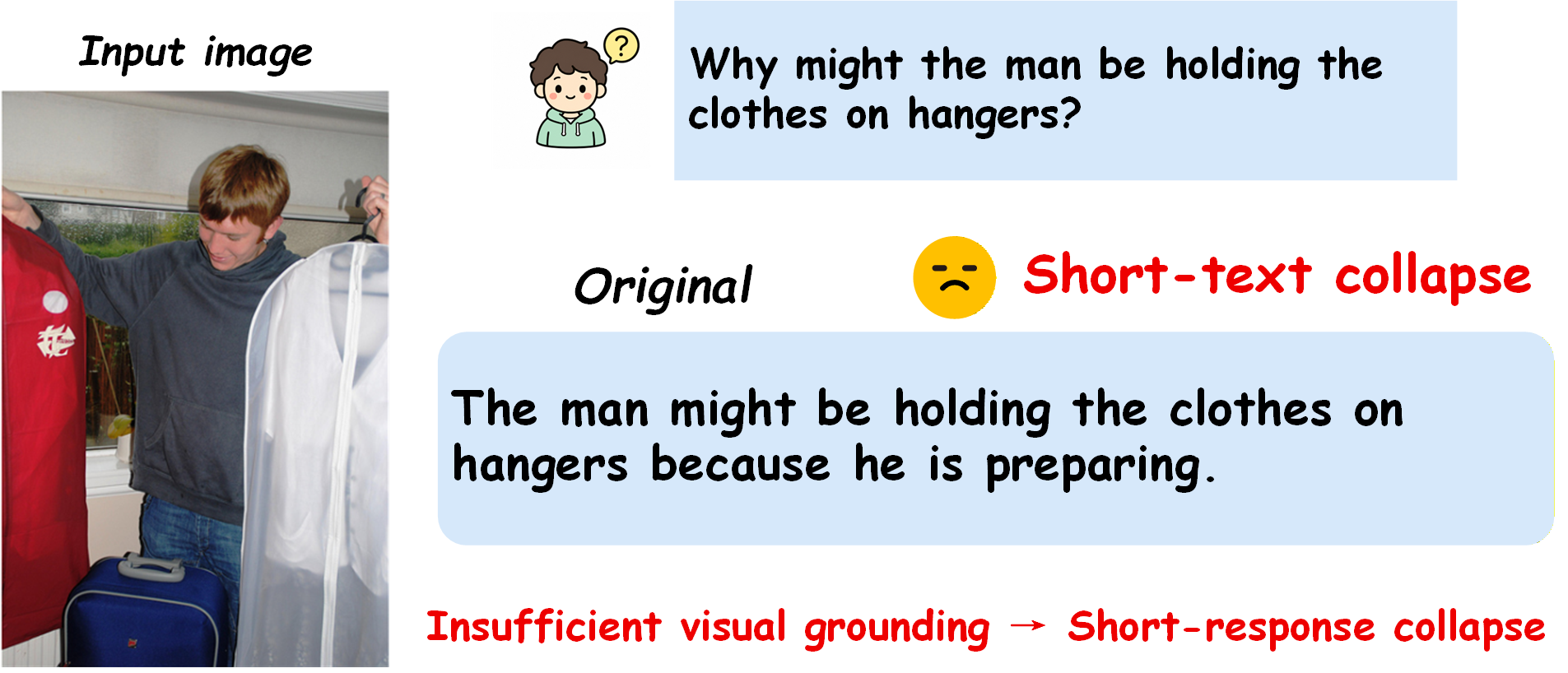}
  \caption{A representative short-response collapse associated with
  insufficient visual grounding.}
  \label{fig:qual_short}
\end{figure}

However, confidence remains an imperfect criterion for commit scheduling. As
illustrated in Figure~\ref{fig:qual_short}, high-confidence candidates may be
driven primarily by the language prior rather than visual evidence, leading to
unsupported early commits that propagate through subsequent denoising steps.
To this end, we perform paired forward-ablation diagnostics on 60 LLaVABench
samples and find that the confidence of 87.5\% of high-confidence commits
\emph{increases} after removing the image, indicating that nearly 90\% are
driven by the language prior rather than visual evidence. Degenerate and
hallucination cases also exhibit higher deep-layer attention entropy (Cohen's
$d=1.79$ and $0.72$), revealing ineffective visual evidence aggregation before
commitment. Consequently, existing confidence-based methods regulate only
\emph{when} or \emph{how many} tokens are committed, but not \emph{whether}
committed candidates are visually grounded. Therefore, this raises a
fundamental question: \textit{can commit scheduling distinguish visually
grounded candidates from language-driven ones in a training-free manner?}

In this paper, we propose CD-APC (Confidence-Decomposed Adaptive Parallel
Commitment), which performs evidence-adjusted scoring over a
base-supported pool, determines whether to commit through hierarchical
constraints, and uses an inverse-window mechanism to adapt to inter-step
difficulty. Specifically, CD-APC executes one paired forward pass per step,
comprising a visual branch
forward and an ablation branch forward. It first filters out low-probability
noise within the base-supported pool constructed by the visual branch, then selects a
candidate via evidence-adjusted scoring, and finally confirms the write through a
dual commit gate based on model acceptability and contrastive concentration,
thereby forming a layered constraint of token-level base-support filtering
plus position-level dual gating. To further balance quality and efficiency,
we introduce the CCAW, an inverse-window strategy, which adaptively adjusts the
parallel commit width according to inter-step decoding difficulty. Extensive
experiments show that the proposed framework
improves the reliability of commit scheduling while reducing the decoding
steps to 40--60\% of standard decoding without introducing additional
parameters or retraining. Our main contributions are as follows:
\begin{itemize}
    \item We reveal a disconnect issue between \dMLLM{} confidence and visual
    evidence, and quantify the degraded attention organization underlying two
    representative failure modes.

    \item We propose CD-APC, a hierarchical commit mechanism in which a
    base-supported pool removes spurious contrastive signals from
    low-probability regions, while dual gating ensures that committed candidates
    are supported by both base probability and concentrated visual contrastive
    evidence.

    \item We design CCAW, an inverse-window scheduling strategy that dynamically
    adjusts the parallel width according to inter-step difficulty, balancing
    commit quality and decoding efficiency.
\end{itemize}


\section{Background and Related Work}
\label{sec:related_work}

\subsection{Diffusion Language and Multimodal Models}
\label{sec:rw_dllm}

Masked diffusion language models (dLLMs) start from a fully masked sequence,
predict all masked positions at each step, and selectively retain a subset of
candidates while remasking the rest for later reconsideration
\citep{austin2021d3pm,sahoo2024mdlm,lou2024sedd,shi2024md4,ou2024radd,nie2024smdm}.
LLaDA, Dream, and blockwise diffusion models scale this paradigm to strong
general-purpose language models
\citep{nie2025llada,bie2025llada2,ye2025dream,song2025seed,arriola2025blockwise}; MMaDA, LLaDA-V, LaViDa,
and Dimple further extend it to image-conditioned multimodal decoders
\citep{yang2025mmada,you2025lladav,li2025lavida,yu2025dimple,
swerdlow2025unified,ye2025dreamvl,zeng2025diffusionvl}.
In parallel denoising, early visual predictions can influence many subsequent
positions, making the remasking choice a critical intervention point for visual
evidence fusion.

\subsection{Commit Scheduling}
\label{sec:rw_confidence}

Mainstream commit scheduling is based on a single confidence score $c_{t,i} = p^I_{t,i}(\hat{y}_{t,i})$. Subsequent improvements fall into two categories
\citep{kang2025parallelbench,wei2025slowfast,wang2025d2f,xiang2026cover}.
\textbf{Adaptive budget and deferred commitment}: Fast-dLLM and Dimple propose model-adaptive acceptance counts
\citep{wu2025fastdllm,yu2025dimple,wu2025fastdllmv2,jiang2025d2cache,
huang2025maskprophet}. Later work targets complementary failure
modes: AdaBlock-dLLM adapts the eligible semantic region
\citep{lu2026adablock,zhang2026swordsman}, DCD keeps uncertain positions masked across block
boundaries so that more context can arrive \citep{shu2026dcd}, CCD adds cross-step
predictive consistency \citep{chen2026ccd}, and Prophet uses global answer
convergence to terminate refinement early \citep{li2026prophet}. \textbf{Position--step prior}: Thinking Diffusion introduces
\begin{equation}
\tilde{c}_{t,i} \;=\; c_{t,i}\,\bigl[\,1 - \gamma\,(1 - \tau_t)\,\mathrm{rel}(i)\,\bigr],
\label{eq:psp}
\end{equation}
which discounts confidence by denoising progress and position, deferring later answer positions \citep{kim2026thinking}. All of the above methods revolve around confidence for judging readiness, but none examines whether confidence comes from image support or the language prior.

\subsection{Visual Contrast and Grounded Decoding}
\label{sec:rw_contrast}

Another line of work introduces anchoring signals by comparing predictions with and without image conditioning. VCD uses corrupted-image contrast; M3ID and Visual Debias Decoding use image-free references
\citep{leng2024vcd, favero2024m3id, zhang2024vdd}. CEBaG computes visual sensitivity scores in advance, which are not updated as decoding proceeds \citep{asadi2026cebag}. Thinking Diffusion's VRG modifies denoising logits via
\begin{equation}
\ell^{\mathrm{VRG}}_{t}
\;=\; \ell^{I}_{t} + s\,\bigl(\ell^{I}_{t} - \ell^{\varnothing}_{t}\bigr),
\label{eq:vrg}
\end{equation}
so an image-versus-no-image contrast can indirectly influence the model's commit decisions; the explicit PSP delay, however, still follows the fixed
position-and-step prior above \citep{kim2026thinking}. VISAGE, ST-Veto, and VRCD
use attention weights to gauge each candidate's image dependence and thereby alter commit priority \citep{narnaware2026visage, kim2026stveto, yuan2026vrcd}.
However, attention can only measure whether the model consulted the image, not
whether the image actually raised the model's support for the committed candidate.

None of these works directly uses visual contrast to influence specific
candidate tokens and decides whether to commit. A qualifying scheduler can be
abstracted as
\begin{equation}
\mathcal{S}_t
\;=\; \pi\!\Bigl(\, \bigl\{(c_{t,i},\, g_{t,i},\, h_{t,i})\bigr\}_{i \in \mathcal{M}_t}\Bigr),
\label{eq:unified_scheduler}
\end{equation}
where $c$ measures current readiness, $g$ measures signed visual support, and an
optional $h$ summarizes temporal stability. CD-APC implements this view,
obtaining visual evidence gain through paired forward passes and using it as an
online commit condition. Because $g \approx 0$ leaves candidates unchanged,
neutral function words incur no penalty.


\section{Motivation and Analysis}
\label{sec:motivation}

\subsection{Preliminaries}
\label{sec:mot_pre}

\paragraph{Target model.}
Our diagnostic target is the diffusion multimodal large language model
(\dMLLM). Given image $I$ and text instruction $q$, the model projects both
into a shared token space and performs $N$ parallel denoising steps over a
fully masked response $S_m$. At step $t$, every masked position $i \in
\mathcal{M}_t$ receives a prediction
\begin{equation}
  \pimg(v) \;=\; p_\theta\!\bigl(x_i = v \,\big|\, x^{t}, q, I\bigr).
  \label{eq:mot_pimg}
\end{equation}

\paragraph{Confidence-based scheduling.}
Mainstream \dMLLM{}s use confidence-based remasking. The argmax candidate and
its probability serve as the commitment confidence:
\begin{equation}
  \hat{y}^{I}_{t,i} \;=\; \arg\max_{v}\pimg(v),
  \qquad
  \confimg \;=\; \pimg\!\bigl(\hat{y}^{I}_{t,i}\bigr),
  \label{eq:mot_conf}
\end{equation}
and a scheduler $\pi$ selects $\mathcal{S}_t \subseteq \mathcal{M}_t$ based
on $\confimg$. The same scalar thus decides both \emph{which} candidate is
chosen and \emph{when} it is written back.

\paragraph{Analysis benchmark.}
All diagnostics below use LLaVABench-in-the-wild: 60 samples covering
description, reasoning, and entity recognition.

\subsection{Failure Modes of Confidence-Based Decoding}
\label{sec:mot_failure}

\paragraph{Failure profile.}
Standard confidence decoding on LLaVABench produces two typical failures:
\textbf{short-text degeneration}: output collapses into fragments
such as ``\texttt{5}'', ``\texttt{Jim}'', or ``\texttt{Mona}'', or
enters a repetition loop \citep{zhao2026anchors}; and
\textbf{long-text hallucination}: output
length and fluency are normal, but visual entities are wrong
\citep{rohrbach2018chair}.
Both point to the same problem: a mismatch between commitment timing and
visual evidence.

\paragraph{Prevalence.}
After GPT-5.6 scoring with human verification, the 60 samples split into: \emph{normal} 20/60,
\emph{degenerate} 19/60, \emph{long-text hallucination} 12/60, borderline
9/60. Under the standard threshold, $51.7\%$ fall into the two failure modes.

\medskip
\noindent
\textbf{Finding 1.} Confidence-based scheduling consistently produces
short-text degeneration and long-text hallucination in \dMLLM{} multimodal
generation. These are not attributable to model capacity but to a deficiency of confidence itself.

\subsection{Root Causes: Mismatch between Confidence and Visual Evidence}
\label{sec:mot_root}

\subsubsection{High confidence does not imply image dependence.}
\label{sec:mot_conf_vs_gain}

To quantify each candidate's dependence on the image, we run a paired forward
pass before every commitment. The normal branch reads image tokens; the
ablation branch blocks the output-to-image path via attention bias, with all
other context identical. We define the visual evidence gain as
\begin{equation}
  \gvis(v) \;=\; \log \pimg(v) \;-\; \log \pabl(v),
  \label{eq:mot_gain}
\end{equation}
where $\pabl$ is the ablated-branch distribution. $\gvis$ measures the marginal
impact of image access: positive means the image supports the candidate;
negative means suppression.

\begin{figure}[tb]
  \centering
  \includegraphics[width=0.98\linewidth]{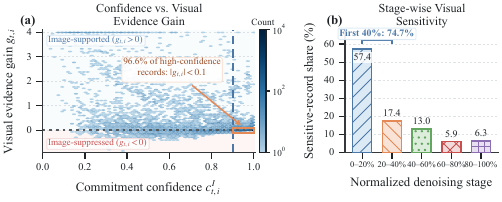}
  \caption{Confidence does not reliably indicate visual support.
  (a) Most high-confidence commits have near-zero visual gain; dashed and
  dotted lines mark $c=0.9$ and $g=0$.
  (b) The first $40\%$ contains $74.7\%$ of visually
  sensitive records.}
  \label{fig:mot_conf_gain}
\end{figure}

\paragraph{Joint distribution.}
Figure~\ref{fig:mot_conf_gain}(a) plots $15{,}360$ records. Of all candidates,
$76.0\%$ exceed the $0.9$ threshold under the normal forward pass; among these
high-confidence candidates, $96.6\%$ satisfy $|\gvis| < 0.1$ and $87.5\%$
satisfy $\gvis \le 0$. Nearly $90\%$ high-confidence commitments are
insensitive to image access; some even gain probability under visual ablation.

\paragraph{Temporal structure.}
Figure~\ref{fig:mot_conf_gain}(b) and Table~\ref{tab:mot_stage_share} show that
visually sensitive positions concentrate in early denoising. $57.4\%$ of
records with $\gvis > 0.5$ appear in the first $20\%$ of the schedule and
$74.7\%$ by $40\%$. In the second half, language priors dominate as
$\gvis$ approaches zero.

\begin{table}[tb]
  \centering
  \small
  \setlength{\tabcolsep}{4pt}
  \begin{tabular}{lcc}
    \toprule
    Denoising stage & Stage share (\%) & In-stage $\gvis{>}0.5$ (\%) \\
    \midrule
    0--20\%    & 57.4 & 34.8 \\
    20--40\%   & 17.4 & 14.0 \\
    40--60\%   & 13.0 & \phantom{0}7.1 \\
    60--80\%   & \phantom{0}5.9 & \phantom{0}2.8 \\
    80--100\%  & \phantom{0}6.3 & \phantom{0}1.2 \\
    \bottomrule
  \end{tabular}
  \caption{Distribution of visually sensitive records ($\gvis > 0.5$) across
  normalized denoising stages.}
  \label{tab:mot_stage_share}
\end{table}

\medskip
\noindent
\textbf{Finding 2.} High confidence does not imply image dependence. Visually
sensitive positions are sparse and front-loaded; middle-to-late
high-confidence commits are driven by the language prior. Confidence cannot
distinguish image-supported from prior-supported candidates.

\subsubsection{Attention organization degrades in failure samples.}
\label{sec:mot_attn}

Total attention mass cannot be directly equated with ``looking more or less
at the image.'' We instead measure the \emph{normalized attention entropy
within the image region}:
\begin{equation}
\begin{aligned}
  \Himg_{l,t,i}
  &\;=\;
  -\,\frac{1}{\log|\mathcal{V}|}
  \sum_{j\in\mathcal{V}}
  \tilde{A}^{\,l,t}_{i,j}\,\log\tilde{A}^{\,l,t}_{i,j}, \\
  \tilde{A}^{\,l,t}_{i,j}
  &\;=\;
  \frac{A^{\,l,t}_{i,j}}{\sum_{k\in\mathcal{V}}A^{\,l,t}_{i,k}},
\end{aligned}
\label{eq:mot_Himg}
\end{equation}
where $\mathcal{V}$ is the image-token interval and $A^{\,l,t}_{i,j}$ is the
head-averaged attention weight from position $i$ to key $j$ at layer $l$,
step $t$. Higher $\Himg$ means more dispersed attention over image
tokens. Statistics are aggregated within each
sample across steps and sampled layers; samples serve as independent
observations for between-group tests.

\begin{figure}[tb]
  \centering
  \includegraphics[width=0.98\linewidth]{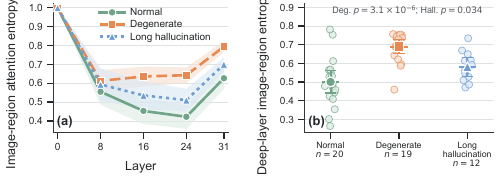}
  \caption{Failure cases exhibit more dispersed image-region attention
  than normal outputs. (a) Cross-layer normalized entropy.
  (b) Deep-layer entropy over layers $16$, $24$, and $31$; error bars show
  $95\%$ confidence intervals, and $p$ values are from two-sided Welch tests.}
  \label{fig:mot_attn_org}
\end{figure}

\paragraph{Cross-layer curves.}
In Fig.~\ref{fig:mot_attn_org}(a), all groups begin with high entropy in
shallow layers. Past $l \ge 8$, Normal samples drop rapidly and stabilize at
$0.42$--$0.55$; degenerate samples barely decline; long-text hallucination
falls in between.

\begin{figure}[tb]
  \centering
  \includegraphics[width=\linewidth]{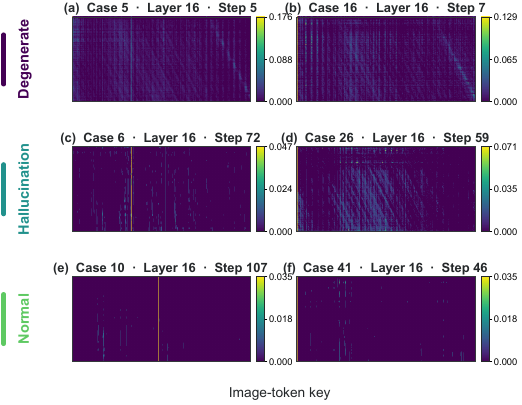}
  \caption{Output-to-image attention at layer $16$. Degenerate and
  hallucinated outputs show dispersed or drifting patterns, whereas normal
  outputs retain sparse visual anchors.}
  \label{fig:mot_attn_qual}
\end{figure}

\paragraph{Qualitative comparison.}
Fig.~\ref{fig:mot_attn_qual} compares two samples per group rather than a
single selected case. Normal samples preserve sparse vertical anchors across
output positions, indicating stable attention to a small set of image-token
regions. Degenerate samples distribute weak attention across many keys or
collapse onto isolated columns, with little coherent spatial structure.
Long-text hallucinations combine persistent narrow-column fixation with
diagonal drift, consistent with wrong-region focus or output self-loops.

\medskip
\noindent
\textbf{Finding 3.} Both failure groups show significant degradation of
deep-layer image-attention \emph{organization}, but in distinct forms:
degeneration is dominated by strong within-region dispersion; long-text
hallucination shows moderate dispersion plus wrong-region focus or
self-loops. Although total attention mass is similar across groups, attention
\emph{organization} degrades significantly in the failure groups.

\subsection{Discussion and Design Implications}
\label{sec:mot_discussion}

Findings~1 and~2 show that confidence admits both visually grounded and
language-prior-driven candidates without distinguishing their source. Yet
introducing visual contrast alone does not solve the problem. Therefore, visual
contrast must be restricted to the model-approved candidate space and
commitment must be gated.

Finding~3 further reveals
significant variation across samples in the ability to form stable visual
organization. A fixed parallel width either under-scrutinizes tokens in
high-risk steps or wastes computation in low-risk steps. Parallel width
should therefore adapt to per-step commit risk.

Accordingly, we propose Confidence-Decomposed Adaptive Parallel Commitment
(CD-APC), combining a base-supported evidence-adjusted dual-gated commit
mechanism with an inverse window: the former improves commit quality; the
latter dynamically regulates commit capacity.


\begin{figure*}[t]
  \centering
  \includegraphics[width=0.96\linewidth]{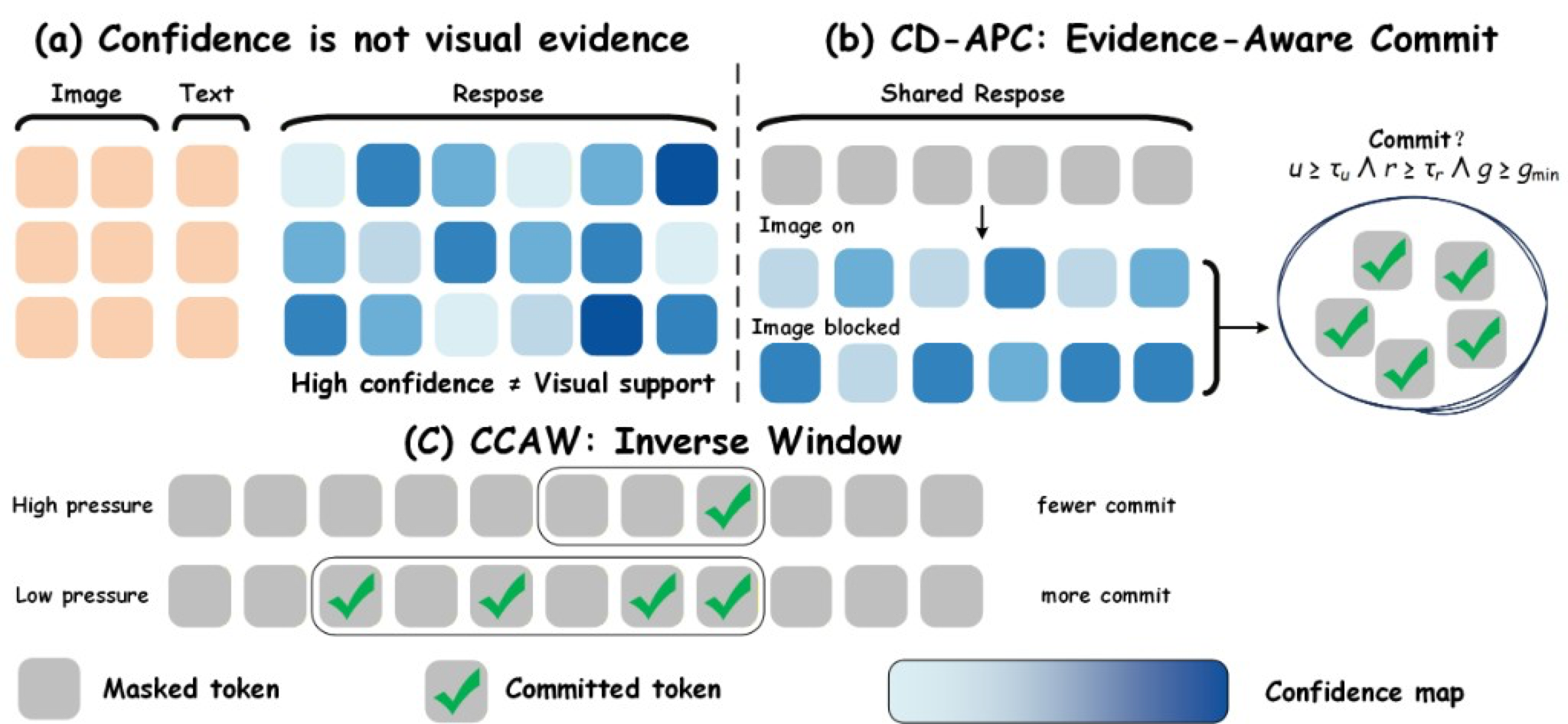}
  \caption{Overview of CD-APC. (a)~Confidence alone cannot identify visual
  support. (b)~Evidence-adjusted scoring and dual gating filter unsupported
  commits. (c)~CCAW adapts the commit window to step pressure.}
  \label{fig:overview}
\end{figure*}

\section{Method}
\label{sec:method}

Section~\ref{sec:motivation} shows that the core deficiency of standard decoding is that confidence does not distinguish whether a candidate comes from the image or the language prior. CD-APC departs from this deficiency: it first performs evidence-adjusted scoring within the base-supported pool, then subjects commitment to dual gating controlled by model acceptability and contrastive concentration, ensuring that every candidate written into the sequence has both baseline probability support and a concentrated contrastive signal. On this basis, an inverse window adapts to inter-step difficulty variation.

\subsection{Paired Forward and Base-Supported Pool}
\label{sec:meth_paired}

At each denoising step $t$, a paired forward pass is executed for all pending positions. The visual branch retains the full image input, producing distribution $p^I_{t,i}$; the ablation branch blocks text-to-image-key attention paths, producing $p^A_{t,i}$. Both branches share the sequence state and are merged into a single batch-2 model call.

The search space for contrast should not cover the full vocabulary because low-probability tokens may receive spuriously high evidence-adjusted scores due to extreme suppression in the ablation branch. We therefore measure each token's support relative to the visual-branch top-1:
\begin{equation}
  \delta_{t,i}(v)
  =
  \log p^I_{t,i}(v)
  -
  \max_w\log p^I_{t,i}(w).
  \label{eq:base_pool}
\end{equation}
With $\beta=0.1$, tokens satisfying $\delta_{t,i}(v)\ge\log\beta$ form the base-supported pool. Using the visual evidence gain in Eq.~\eqref{eq:mot_gain}, we define the evidence-adjusted score and candidate as
\begin{equation}
  \begin{aligned}
  h_{t,i}(v)
  &=
  \begin{cases}
  \log p^I_{t,i}(v)+\alpha g_{t,i}(v),
    & \delta_{t,i}(v)\ge\log\beta,\\
  -\infty, & \text{otherwise},
  \end{cases}\\
  \hat y^c_{t,i}
  &=\arg\max_v h_{t,i}(v).
  \end{aligned}
  \label{eq:score}
\end{equation}

\subsection{Dual-Gated Commit Criterion}
\label{sec:meth_commit}

The base-supported pool restricts contrast to the model-approved candidate space, eliminating spurious contrastive signals in low-probability regions. On this basis, a candidate must pass two complementary gating checks before being written into the sequence:

\paragraph{Acceptability gate ($u$).}
$u = p^I(\hat{y}^c)$: the probability of the candidate under the full visual
model. Evidence-adjusted scoring may promote a token other than the original top-1
as the best candidate, but such a correction is not credible if the model
assigns it almost zero probability. $u$ keeps the candidate within
the model's plausible range.

\paragraph{Contrastive concentration gate ($r$).}
Let $q_{t,i}=\mathrm{softmax}(h_{t,i})$ be the score-normalized distribution and $m_{t,i}=\sum_{v:\delta_{t,i}(v)\ge\log\beta}p^I_{t,i}(v)$ the pool mass:
\begin{equation}
  r_{t,i} = m_{t,i}\,q_{t,i}(\hat{y}^c_{t,i}).
  \label{eq:readiness}
\end{equation}
$q(\hat y^c)$ reflects the candidate's concentration in the proposal ranking; $m$ penalizes degenerate cases where the candidate pool carries only a small fraction of total probability. When the proposal distribution vacillates among multiple candidates, $r$ is low, preventing premature commitment.

\medskip
\noindent The full commit condition is:
\begin{equation}
\boxed{\;
  \mathrm{commit}(t,i)
  \;\Longleftrightarrow\;
  u\ge\tau_u
  \;\land\;
  r\ge\tau_r
\;}
  \label{eq:commit}
\end{equation}
We set $\tau_u{=}0.10$, $\tau_r{=}0.90$. The two thresholds operate at different scales and serve complementary roles. $\tau_u$ is a permissive lower bound: if contrast does flip the top-1, then $u < p^I(\hat y^I)$; setting $\tau_u$ too high would by definition prohibit all corrections. $\tau_r$ is a strict concentration requirement, ensuring commitment occurs only when the contrastive signal is highly concentrated on a single candidate.

\paragraph{Degenerate handling.}
When the base-supported pool contains one token, $q(\hat y^c){=}1$ and $r{=}u$; hence
Eq.~\eqref{eq:commit} reduces to
$u\ge\max(\tau_u,\tau_r)=0.90$ under our thresholds.

\paragraph{Progress guarantee.}
For position $i$, define the normalized joint readiness
$\rho_i=\min(u_i/\tau_u,\,r_i/\tau_r)$. If no position in the window satisfies
Eq.~\eqref{eq:commit}, the position with highest $\rho_i$ commits its original
top-1 (not the contrastive candidate). This path does not use contrastive
results and solely ensures denoising does not deadlock.

\subsection{Inverse-Window Adaptation (CCAW)}
\label{sec:meth_ccaw}

Window capacity $W_t$ determines how many pending positions enter the
left-anchored candidate set at step $t$; the separate budget $B$ limits how
many eligible positions are committed. A fixed window cannot track
inter-step difficulty. CCAW adapts inversely to step pressure:
\begin{gather}
  d_t = \tfrac{1}{3}(f_{\mathrm{conf}}+f_{\mathrm{def}}+f_{\mathrm{inst}}),
  \label{eq:ccaw_d}\\
  W^* = W_{\max} - \lfloor(W_{\max}{-}W_{\min})\cdot d_t\rceil.
  \label{eq:ccaw}
\end{gather}
$f_{\mathrm{conf}}$ is the fraction of positions where contrast flipped the
candidate and $\rho_i\ge1$, meaning that both gates meet their thresholds;
$f_{\mathrm{def}}$ is the fraction failing Eq.~\eqref{eq:commit};
$f_{\mathrm{inst}}$ is the fraction with inconsistent candidates across
consecutive steps. Actual capacity approaches $W^*$ with asymmetric step sizes
(shrink~8, expand~4) and EMA smoothing. Under high pressure the window
contracts for narrower candidate selection; under low pressure it expands for
broader batch selection, reducing decoding steps to 40--60\% of the baseline.

\begin{table*}[t]
  \centering
  \small
\setlength{\tabcolsep}{1.8pt}
\begin{tabular*}{\textwidth}{@{\extracolsep{\fill}}ll cc cc cc cc cc@{}}
    \toprule
    \multirow{2}{*}{\textbf{Model}}
      & \multirow{2}{*}{\textbf{Method}}
      & \multicolumn{2}{c}{\textbf{LLaVA} \textit{512/256}}
      & \multicolumn{2}{c}{\textbf{MMMU} \textit{128/64}}
      & \multicolumn{2}{c}{\textbf{MMBench} \textit{128/64}}
      & \multicolumn{2}{c}{\textbf{M3CoT} \textit{512/256}}
      & \multicolumn{2}{c}{\textbf{VLind} \textit{128/64}} \\
    \cmidrule(lr){3-4}\cmidrule(lr){5-6}\cmidrule(lr){7-8}
    \cmidrule(lr){9-10}\cmidrule(lr){11-12}
      & & Sc$\uparrow$ & s/it$\downarrow$
      & Acc$\uparrow$ & s/it$\downarrow$
      & Acc$\uparrow$ & s/it$\downarrow$
      & Acc$\uparrow$ & s/it$\downarrow$
      & $S_{\mathrm{LP}}\uparrow$ & s/sample$\downarrow$ \\
    \midrule
    \multicolumn{12}{l}{\textit{\textbf{MMaDA-8B-MixCoT}}} \\
      & Original
        & 12.9 & 12.1 & 30.7 & 29.6 & 55.8 & 10.5
        & 36.0 & 29.7 & \phantom{0}9.9 & 211.9 \\
      & DCD
        & \textbf{32.2} & 2.7 & 30.7 & 1.0
        & 56.3 & 6.2
        & 37.3 & 2.4 & \phantom{0}1.7 & 8.6 \\
      & VCD
        & 15.3 & 20.9 & \underline{31.0} & 20.6
        & \underline{56.4} & 22.5
        & \underline{37.6} & 29.4 & 20.8 & 409.5 \\
      & SWD
        & 18.1 & 13.6 & 30.4 & 12.3 & 55.4 & 11.6
        & 35.2 & 16.5 & \phantom{0}9.3 & 293.1 \\
      & PSP
        & 16.5 & 12.3 & 30.6 & 11.0 & 55.4 & 10.5
        & 36.4 & 14.9 & 19.6 & 335.8 \\
      & PSP+VRG
        & 19.0 & 38.4 & 29.8 & 19.9 & 56.1 & 18.7
        & 36.9 & 26.8 & \textbf{25.0} & 877.0 \\
      & \textbf{CD-APC}
        & \underline{32.1} & 13.4 & \textbf{31.5} & 2.5
        & \textbf{56.9} & 2.3
        & \textbf{38.6} & 9.0 & \underline{21.1} & 52.3 \\
    \midrule
    \multicolumn{12}{l}{\textit{\textbf{LaViDa-Reason}}} \\
      & Original
        & 54.5 & 4.3 & 33.8 & 3.9 & 71.7 & 3.9
        & 45.1 & 11.8 & 52.9 & 15.3 \\
      & DCD
        & \underline{73.4} & 5.1 & 32.5 & 3.7 & 73.4 & 3.7
        & 31.5 & 7.2 & 50.0 & 76.2 \\
      & VCD
        & 64.8 & 8.4 & \underline{36.0} & 5.4 & 71.0 & 5.2
        & 45.2 & 24.1 & \underline{53.9} & 130.4 \\
      & SWD
        & 60.4 & 4.0 & 33.5 & 3.8 & 71.2 & 3.6
        & \textbf{45.5} & 11.2 & 44.7 & 73.0 \\
      & PSP
        & 56.3 & 4.1 & 35.9 & 3.8 & 73.2 & 3.6
        & \underline{45.4} & 11.2 & 32.5 & 79.7 \\
      & PSP+VRG
        & 59.8 & 7.7 & 35.8 & 5.3
        & \textbf{74.1} & 4.9
        & \textbf{45.5} & 33.3 & \textbf{57.8} & 118.2 \\
      & \textbf{CD-APC}
        & \textbf{74.6} & 7.9 & \textbf{36.1} & 4.3
        & \underline{73.9} & 3.8
        & \underline{45.4} & 22.9 & \textbf{57.8} & 65.9 \\
    \bottomrule
\end{tabular*}
  \caption{Main results across two dMLLM backbones (within-model
  comparisons). Headers show max new tokens/denoising steps.
  \textbf{Bold}: best quality score per backbone and metric.
  \underline{Underline}: second best. Dashes denote unreported results.}
  \label{tab:main}
\end{table*}


\begin{algorithm}[t]
\caption{CD-APC}
\label{alg:cdapc}
\small
\begin{algorithmic}[1]
\REQUIRE Model $M$, fully-masked sequence $x$, image span $[a,b)$, commit budget $B$
\STATE Initialize $W_1\leftarrow W_{\min}$
\WHILE{$\exists\,i: x_i=\texttt{MASK}$}
  \STATE $\mathcal{M}_t\leftarrow\{i:x_i=\texttt{MASK}\}$
  \STATE $\mathcal{C}_t\leftarrow\operatorname{Leftmost}_{W_t}(\mathcal{M}_t)$
  \hfill\COMMENT{candidate window}
  \STATE $(p^I, p^A)\leftarrow$ paired\_forward$(M,x,[a,b))$ \hfill\COMMENT{batch-2}
  \STATE For $i\in\mathcal{C}_t$, compute
  $\delta_i,h_i,\hat{y}^c_i,u_i,r_i$
  \STATE $\mathcal{E}_t\leftarrow
  \{i\in\mathcal{C}_t:u_i\ge\tau_u\land r_i\ge\tau_r\}$
  \IF{$\mathcal{E}_t\neq\emptyset$}
    \STATE $\mathcal{S}_t\leftarrow
    \operatorname{Top}_{\min(B,|\mathcal{E}_t|)}(\mathcal{E}_t;r_i)$
    \STATE Commit $\hat y^c_i$ for all $i\in\mathcal{S}_t$
  \ELSE
    \STATE $j\leftarrow\arg\max_{i\in\mathcal{C}_t}\rho_i$;
    commit original $\hat y^I_j$
  \ENDIF
  \STATE Compute $d_t$; update $W_{t+1}$ toward $W^*$ using Eq.~\eqref{eq:ccaw}
\ENDWHILE
\end{algorithmic}
  \noindent Each step requires one paired forward pass.
\end{algorithm}


\section{Experiments}
\label{sec:exp}

\subsection{Setup}
\label{sec:exp_setup}

\paragraph{Model and baselines}
All experiments are based on MMaDA-8B-MixCoT~\citep{yang2025mmada} and
LaViDa-Reason~\citep{li2025lavida}, two 8B-parameter diffusion multimodal
large language models. The baselines
include: (1)~\textbf{Original}, which uses standard fixed-step blockwise
diffusion; (2)~\textbf{VCD}, which contrasts predictions from original and
distorted images under a plausibility constraint~\citep{leng2024vcd};
(3)~\textbf{DCD}, a
confidence-aware deferred-commitment method for text-only models that we
extend to the multimodal setting~\citep{shu2026dcd}; (4)~\textbf{SWD},
which measures the temporal instability of a token using the KL divergence
between prediction distributions at adjacent denoising steps and reweights
confidence accordingly~\citep{wu2026swd}; (5)~\textbf{PSP}, which discounts
the confidence according to denoising progress and the token's relative
position~\citep{kim2026thinking}; and (6)~\textbf{PSP+VRG}, which adds visual
reasoning guidance to PSP~\citep{kim2026thinking}.

\paragraph{Benchmarks}
We evaluate on five benchmarks spanning open-ended generation,
multi-discipline reasoning, general multimodal understanding, multi-step
chain-of-thought, and visual--language grounding: (1)~LLaVABench
~\citep{liu2023llava} for open-ended generation quality;
(2)~MMMU~\citep{yue2024mmmu} for multi-discipline
multiple-choice accuracy; (3)~MMBench~\citep{liu2024mmbench} for
multimodal understanding accuracy; (4)~M3CoT~\citep{chen2024m3cot} for
multimodal multi-step chain-of-thought reasoning; and
(5)~VLind-Bench~\citep{lee2024vlind}, a language-prior diagnostic benchmark
using pipeline evaluation.

\subsection{Main Results}
\label{sec:exp_main}

Table~\ref{tab:main} tests whether visual-evidence-aware commitment transfers
across two dMLLM backbones with different decoding dynamics.  On MMaDA,
CD-APC improves Original by 19.2 LLaVA points and by
0.9, 1.2, 2.6, and 11.2\,pp on MMMU, MMBench, M3CoT, and VLind,
respectively.  On LaViDa, CD-APC improves LLaVA, MMMU, and MMBench by
20.1, 2.3, and 2.2\,pp, respectively.  These gains show
that conditioning commitment on visual evidence is not tied to a single
model or confidence landscape.

\paragraph{Grounded understanding and multi-step reasoning.}
On MMaDA, CD-APC obtains the best MMMU (31.5\%), MMBench (56.9\%), and
M3CoT (38.6\%) results. On LaViDa, CD-APC obtains the best MMMU accuracy
(36.1\%) and reaches 73.9\% on MMBench, while attaining the strongest LLaVA
score among the methods in Table~\ref{tab:main}. The M3CoT gain on MMaDA
(+2.6\,pp) is especially informative as visual evidence must be carried
across multiple reasoning steps, selecting an image-supported candidate
prevents an early unsupported commit from propagating downstream.

\paragraph{Language-prior robustness}
VLind-Bench isolates whether a model answers from the image or from a language
prior. On MMaDA, CD-APC more than doubles $S_{\mathrm{LP}}$ from 9.9\% to
21.1\%, whereas DCD records 1.7\% despite its strong LLaVABench score. PSP+VRG
reaches 25.0\% by applying full visual guidance, while CD-APC reaches 21.1\%
at 52.3\,s/sample rather than 877.0\,s/sample. Thus, our method retains a
strong visual-grounding signal at a practical inference cost. Supplementary
results provide the complete CK/VP/CB/LP breakdown.

\paragraph{Generation quality and grounding}
CD-APC raises the MMaDA LLaVABench score from 12.9 to 32.1, essentially
matching DCD at 32.2.  Their VLind scores, however, separate the mechanisms:
21.1 for CD-APC versus 1.7 for DCD.  This controlled comparison isolates the
benefit of the visual gate beyond deferred commitment: both schedules resolve
short-text collapse, but evidence-aware commitment also preserves image
dependence.  On LaViDa LLaVA-COCO, CD-APC raises the overall score from
54.5 to 74.6 while requiring 7.9\,s/it.

\paragraph{Efficiency on MMaDA}
The quality gains do not require a longer denoising schedule.  CD-APC reduces
wall time from 29.6 to 2.5\,s/sample on MMMU, from 10.5 to 2.3\,s/sample on
MMBench, and from 29.7 to 9.0\,s/sample on M3CoT.  CCAW contracts the
commitment window only when the step is difficult and expands it once the
candidate set stabilizes, concentrating paired-forward computation where
visual scrutiny is most useful.

\subsection{Ablation Study}
\label{sec:exp_ablation}

\begin{table}[tb]
  \centering
  \small
  \setlength{\tabcolsep}{2.0pt}
  \begin{tabular}{@{}lccccc@{}}
    \toprule
    \textbf{Configuration}
      & \textbf{LLaVA}
      & \textbf{MMMU}
      & \textbf{MMB}
      & \textbf{M3CoT}
      & \textbf{VLind} \\
    \midrule
    \multicolumn{6}{l}{\textbf{MMaDA-8B-MixCoT}} \\
    Original
      & 12.9 & \underline{30.7} & 55.8
      & 36.0 & \phantom{0}9.9 \\
    CAFW
      & \underline{31.3} & 30.5 & \underline{56.6}
      & \textbf{38.7} & \underline{20.3} \\
    w/o $u$-gate
      & 30.3 & 29.8 & 44.9
      & \underline{38.6} & 19.3 \\
    w/o $r$-gate
      & 30.5 & 29.8 & 45.0
      & 34.1 & 18.3 \\
    \textbf{CD-APC}
      & \textbf{32.1} & \textbf{31.5}
      & \textbf{56.9} & \underline{38.6}
      & \textbf{21.1} \\
    \midrule
    \multicolumn{6}{l}{\textbf{LaViDa-Reason}} \\
    Original
      & 54.5 & 33.8 & 71.7
      & \underline{45.1} & 52.9 \\
    CAFW
      & \textbf{80.3} & \underline{35.9}
      & \underline{72.1} & 43.4
      & \underline{55.0} \\
    \textbf{CD-APC}
      & \underline{74.6} & \textbf{36.1}
      & \textbf{73.9} & \textbf{45.4}
      & \textbf{57.8} \\
    \bottomrule
  \end{tabular}
  \caption{Component ablation across both backbones. CAFW denotes CD-APC
  with a fixed window.}
  \label{tab:component_ablation}
\end{table}

\paragraph{CAFW (CD-APC fixed window)}
On MMaDA, CAFW recovers generation quality (+18.4 LLaVA), improves M3CoT by
+2.8\,pp, and lifts VLind $S_{\mathrm{LP}}$ by +10.4\,pp. On LaViDa, CAFW
improves LLaVA, MMMU, MMBench, and VLind, while the inverse window subsequently
recovers M3CoT. These results isolate the contribution of visual-evidence-aware
commitment before capacity adaptation.

\paragraph{CD-APC: adding the CCAW inverse window}
On MMaDA, CCAW raises MMMU and MMBench by a further 1.0 and 0.3\,pp,
raises VLind by 0.8\,pp, and reduces wall time from 9.3 to 9.0\,s while
maintaining M3CoT.  On LaViDa, CCAW raises MMMU, MMBench, and VLind by 0.2,
1.8, and 2.8\,pp and reduces LLaVA time from 9.0 to 7.9\,s/it, trading some
open-ended generation score for higher understanding accuracy and lower
latency.

\begin{figure}[t]
  \centering
  \includegraphics[width=\columnwidth]{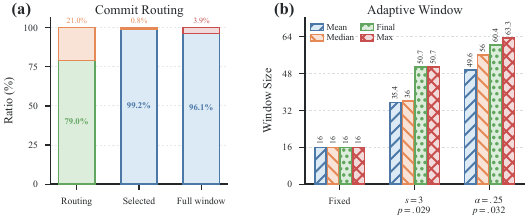}
  \caption{Commit routing and adaptive-window statistics.
  (a) Gated/fallback routing and candidate flip rates.
  (b) Mean, median, final, and maximum window sizes.}
  \label{fig:commit}
\end{figure}

\subsection{Decoding Behavior Analysis}
\label{sec:exp_commit}

The dual gate routes 202.2 gated versus 53.8 fallback tokens, covering 79.0\%
of commitments. Only 0.8\% of selected candidates change versus 3.9\% over
full windows, showing that visual evidence controls commit eligibility
and timing.

From the fixed window of 16, CCAW reaches a mean/final capacity of 35.4/50.7
at $s=3$; $\alpha=0.25$ further yields a mean/median/maximum of 49.6/56/63.3.
Capacity therefore expands under low pressure but contracts on difficult
steps.

\begin{table}[t]
  \centering
  \small
  \setlength{\tabcolsep}{4.5pt}
  \begin{tabular}{@{}lrrrr@{}}
    \toprule
    \multirow{2}{*}{\textbf{Method}}
      & \multicolumn{2}{c}{\textbf{MMaDA}}
      & \multicolumn{2}{c}{\textbf{LaViDa}} \\
    \cmidrule(lr){2-3}\cmidrule(lr){4-5}
      & \textbf{Score}$\uparrow$ & \textbf{Len}
      & \textbf{Score}$\uparrow$ & \textbf{Len} \\
    \midrule
    Original  & 12.9          & 1.8   & 54.5          & 73.6 \\
    DCD       & \textbf{32.2} & 83.0  & 73.4          & 104.2 \\
    SWD       & 18.1          & 9.1   & 60.4          & 37.8 \\
    PSP       & 16.5          & 5.7   & 56.3          & 37.1 \\
    PSP+VRG   & 19.0          & 9.2   & 59.8          & 34.4 \\
    VCD       & 15.3          & 4.3   & 64.8          & 51.2 \\
    CAFW      & 31.3          & 73.0  & \textbf{80.3} & 156.4 \\
    CD-APC    & \underline{32.1} & 70.3  & \underline{74.6} & 95.7 \\
    \bottomrule
  \end{tabular}
  \caption{LLaVABench score and average output length.
  Reporting both distinguishes quality gains from length expansion.
  \textbf{Bold}: best score; \underline{Underline}: second best per model.}
  \label{tab:length_stats}
\end{table}

\paragraph{Recovering non-collapsed outputs.}
Table~\ref{tab:length_stats} shows that Original on MMaDA produces only
1.8 tokens on average. CAFW and CD-APC restore generation to means of 73.0
and 70.3 tokens, while CD-APC remains below DCD's 83.0. On LaViDa, CCAW
reduces the mean length from CAFW's 156.4 to 95.7 tokens, closer to Original's
73.6 and below DCD's 104.2. This shows that CD-APC recovers non-collapsed
generation while controlling length expansion across both backbones.


\section{Conclusion}
\label{sec:conclusion}

We show that confidence magnitude and visual support are distinct signals:
reliable diffusion multimodal decoding requires assessing both confidence and
visual grounding. CD-APC integrates evidence-adjusted scoring over a
base-supported pool, hierarchical commit gating, and an inverse-window strategy
that adapts commit capacity to decoding difficulty.
Experiments on MMaDA-8B-MixCoT and LaViDa-Reason across five benchmarks
demonstrate that CD-APC consistently improves over the Original
decoder, with the largest gains on visual-evidence-sensitive reasoning and
language-prior diagnostic tasks.


\bibliography{references}

@inproceedings{austin2021d3pm,
  title     = {Structured Denoising Diffusion Models in Discrete State-Spaces},
  author    = {Austin, Jacob and Johnson, Daniel D. and Ho, Jonathan and Tarlow, Daniel and van den Berg, Rianne},
  booktitle = {Advances in Neural Information Processing Systems (NeurIPS)},
  volume    = {34},
  pages     = {17981--17993},
  year      = {2021}
}

@inproceedings{sahoo2024mdlm,
  title     = {Simple and Effective Masked Diffusion Language Models},
  author    = {Sahoo, Subham Sekhar and Arriola, Marianne and Schiff, Yair and Gokaslan, Aaron and Marroquin, Edgar and Chiu, Justin T. and Rush, Alexander M. and Kuleshov, Volodymyr},
  booktitle = {Advances in Neural Information Processing Systems (NeurIPS)},
  volume    = {37},
  pages     = {130136--130184},
  doi       = {10.52202/079017-4135},
  year      = {2024}
}

@misc{nie2025llada,
  title         = {Large Language Diffusion Models},
  author        = {Nie, Shen and Zhu, Fengqi and You, Zebin and Zhang, Xiaolu and Ou, Jingyang and Hu, Jun and Zhou, Jun and Lin, Yankai and Wen, Ji-Rong and Li, Chongxuan},
  year          = {2025},
  eprint        = {2502.09992},
  archivePrefix = {arXiv}
}

@misc{ye2025dream,
  title         = {{Dream 7B}: Diffusion Large Language Models},
  author        = {Ye, Jiacheng and Xie, Zhihui and Zheng, Lin and Gao, Jiahui and Wu, Zirui and Jiang, Xin and Li, Zhenguo and Kong, Lingpeng},
  year          = {2025},
  eprint        = {2508.15487},
  archivePrefix = {arXiv}
}

@inproceedings{arriola2025blockwise,
  title     = {Block Diffusion: Interpolating Between Autoregressive and Diffusion Language Models},
  author    = {Arriola, Marianne and Gokaslan, Aaron and Chiu, Justin T. and Yang, Zhihan and Qi, Zhixuan and Han, Jiaqi and Sahoo, Subham Sekhar and Kuleshov, Volodymyr},
  booktitle = {International Conference on Learning Representations (ICLR)},
  volume    = {2025},
  pages     = {50726--50753},
  year      = {2025}
}

@inproceedings{yang2025mmada,
  title     = {{MMaDA}: Multimodal Large Diffusion Language Models},
  author    = {Yang, Ling and Tian, Ye and Li, Bowen and Zhang, Xinchen and Shen, Ke and Tong, Yunhai and Wang, Mengdi},
  booktitle = {Advances in Neural Information Processing Systems (NeurIPS)},
  volume    = {38},
  pages     = {138867--138907},
  year      = {2025}
}

@inproceedings{you2025lladav,
  title     = {{LLaDA-V}: Large Language Diffusion Models with Visual Instruction Tuning},
  author    = {You, Zebin and Nie, Shen and Zhang, Xiaolu and Zhou, Jun and Lu, Zhiwu and Wen, Ji-Rong and Li, Chongxuan},
  booktitle = {Proceedings of the {IEEE/CVF} Conference on Computer Vision and Pattern Recognition (CVPR)},
  pages     = {10093--10105},
  year      = {2026}
}

@inproceedings{li2025lavida,
  title     = {{LaViDa}: A Large Diffusion Language Model for Multimodal Understanding},
  author    = {Li, Shufan and Kallidromitis, Konstantinos and Bansal, Hritik and Gokul, Akash and Kato, Yusuke and Kozuka, Kazuki and Kuen, Jason and Lin, Zhe and Chang, Kai-Wei and Grover, Aditya},
  booktitle = {Advances in Neural Information Processing Systems (NeurIPS)},
  volume    = {38},
  pages     = {105101--105134},
  year      = {2025}
}

@misc{yu2025dimple,
  title         = {Dimple: Discrete Diffusion Multimodal Large Language Model with Parallel Decoding},
  author        = {Yu, Runpeng and Ma, Xinyin and Wang, Xinchao},
  year          = {2025},
  eprint        = {2505.16990},
  archivePrefix = {arXiv}
}

@inproceedings{wu2025fastdllm,
  title     = {{Fast-dLLM}: Training-free Acceleration of Diffusion {LLM} by Enabling {KV} Cache and Parallel Decoding},
  author    = {Wu, Chengyue and Zhang, Hao and Xue, Shuchen and Liu, Zhijian and Diao, Shizhe and Zhu, Ligeng and Luo, Ping and Han, Song and Xie, Enze},
  booktitle = {International Conference on Learning Representations (ICLR)},
  volume    = {2026},
  pages     = {57027--57051},
  year      = {2026}
}

@inproceedings{lu2026adablock,
  title     = {{AdaBlock-dLLM}: Semantic-Aware Diffusion {LLM} Inference via Adaptive Block Size},
  author    = {Lu, Guanxi and Chen, Hao Mark and Karashima, Yuto and Wang, Zhican and Fujiki, Daichi and Fan, Hongxiang},
  booktitle = {International Conference on Learning Representations (ICLR)},
  volume    = {2026},
  pages     = {140018--140033},
  year      = {2026}
}

@misc{shu2026dcd,
  title         = {Deferred Commitment Decoding for Diffusion Language Models},
  author        = {Shu, Yingte and Tian, Yuchuan and Xu, Chao and Wang, Yunhe and Chen, Hanting},
  year          = {2026},
  eprint        = {2601.02076},
  archivePrefix = {arXiv}
}

@misc{wu2026swd,
  title         = {Stability-Weighted Decoding for Diffusion Language Models},
  author        = {Wu, Yue and Huang, Jian},
  year          = {2026},
  eprint        = {2604.17068},
  archivePrefix = {arXiv}
}

@inproceedings{chen2026ccd,
  title     = {Beyond Confidence: Adaptive and Coherent Decoding for Diffusion Language Models},
  author    = {Chen, Kecheng and Liu, Ziru and Tao, Xijia and Liu, Hui and Fu, Xinyu and Zhang, Suiyun and Tu, Dandan and Kong, Lingpeng and Liu, Rui and Li, Haoliang},
  booktitle = {Proceedings of the 43rd International Conference on Machine Learning (ICML)},
  volume    = {306},
  series    = {Proceedings of Machine Learning Research},
  publisher = {PMLR},
  year      = {2026}
}

@inproceedings{li2026prophet,
  title     = {Diffusion Language Model Knows the Answer Before It Decodes},
  author    = {Li, Pengxiang and Zhou, Yefan and Muhtar, Dilxat and Yin, Lu and Yan, Shilin and Shen, Li and Liang, Yi and Vosoughi, Soroush and Liu, Shiwei},
  booktitle = {International Conference on Learning Representations (ICLR)},
  volume    = {2026},
  pages     = {135235--135253},
  year      = {2026}
}

@inproceedings{kim2026thinking,
  title     = {Thinking Diffusion: Penalize and Guide Visual-Grounded Reasoning in Diffusion Multimodal Language Models},
  author    = {Kim, Keuntae and Kang, Mingyu and Choi, Yong Suk},
  booktitle = {Proceedings of the {IEEE/CVF} Conference on Computer Vision and Pattern Recognition (CVPR)},
  pages     = {5154--5164},
  year      = {2026}
}

@inproceedings{kim2026stveto,
  title     = {{ST-Veto}: Spatio-Temporal Token Veto for Diffusion {MLLMs} via Taylor Prediction and Visual Grounding},
  author    = {Kim, Keuntae and Lee, Beomseok and Kim, Hyunwoo and Choi, Yong Suk},
  booktitle = {Proceedings of the 43rd International Conference on Machine Learning (ICML)},
  volume    = {306},
  series    = {Proceedings of Machine Learning Research},
  publisher = {PMLR},
  year      = {2026}
}

@misc{narnaware2026visage,
  title         = {Seeing to Ground: Visual Attention for Hallucination-Resilient {MDLLMs}},
  author        = {Narnaware, Vishal and Gupta, Animesh and Zhai, Kevin and Wang, Zhenyi and Shah, Mubarak},
  year          = {2026},
  eprint        = {2603.25711},
  archivePrefix = {arXiv}
}

@misc{yuan2026vrcd,
  title         = {Visual-Redundancy-Controlled Parallel Decoding for Diffusion-Based Multimodal Large Language Models},
  author        = {Yuan, Yulin and Zhao, Hongshuo and Meng, Xiangming},
  year          = {2026},
  eprint        = {2605.25820},
  archivePrefix = {arXiv}
}

@inproceedings{leng2024vcd,
  title     = {Mitigating Object Hallucinations in Large Vision-Language Models through Visual Contrastive Decoding},
  author    = {Leng, Sicong and Zhang, Hang and Chen, Guanzheng and Li, Xin and Lu, Shijian and Miao, Chunyan and Bing, Lidong},
  booktitle = {Proceedings of the {IEEE/CVF} Conference on Computer Vision and Pattern Recognition (CVPR)},
  pages     = {13872--13882},
  year      = {2024}
}

@inproceedings{favero2024m3id,
  title     = {Multi-Modal Hallucination Control by Visual Information Grounding},
  author    = {Favero, Alessandro and Zancato, Luca and Trager, Matthew and Choudhary, Siddharth and Perera, Pramuditha and Achille, Alessandro and Swaminathan, Ashwin and Soatto, Stefano},
  booktitle = {Proceedings of the {IEEE/CVF} Conference on Computer Vision and Pattern Recognition (CVPR)},
  pages     = {14303--14312},
  year      = {2024}
}

@inproceedings{zhang2024vdd,
  title     = {Debiasing Multimodal Large Language Models via Penalization of Language Priors},
  author    = {Zhang, Yi-Fan and Shi, Yang and Yu, Weichen and Wen, Qingsong and Wang, Xue and Yang, Wenjing and Zhang, Zhang and Wang, Liang and Jin, Rong},
  booktitle = {Proceedings of the {ACM} International Conference on Multimedia ({ACM MM})},
  pages     = {4232--4241},
  doi       = {10.1145/3746027.3755364},
  year      = {2025}
}

@misc{asadi2026cebag,
  title         = {Deterministic Hallucination Detection in Medical {VQA} via Confidence-Evidence {Bayesian} Gain},
  author        = {Asadi, Mohammad and Nedaee, Tahoura and O'Sullivan, Jack W. and Ashley, Euan and Adeli, Ehsan},
  year          = {2026},
  eprint        = {2603.21693},
  archivePrefix = {arXiv}
}

@inproceedings{lou2024sedd,
  title     = {Discrete Diffusion Modeling by Estimating the Ratios of the Data Distribution},
  author    = {Lou, Aaron and Meng, Chenlin and Ermon, Stefano},
  booktitle = {Proceedings of the 41st International Conference on Machine Learning (ICML)},
  volume    = {235},
  series    = {Proceedings of Machine Learning Research},
  pages     = {32819--32848},
  publisher = {PMLR},
  year      = {2024}
}

@inproceedings{shi2024md4,
  title     = {Simplified and Generalized Masked Diffusion for Discrete Data},
  author    = {Shi, Jiaxin and Han, Kehang and Wang, Zhe and Doucet, Arnaud and Titsias, Michalis K.},
  booktitle = {Advances in Neural Information Processing Systems (NeurIPS)},
  volume    = {37},
  pages     = {103131--103167},
  doi       = {10.52202/079017-3277},
  year      = {2024}
}

@inproceedings{ou2024radd,
  title     = {Your Absorbing Discrete Diffusion Secretly Models the Conditional Distributions of Clean Data},
  author    = {Ou, Jingyang and Nie, Shen and Xue, Kaiwen and Zhu, Fengqi and Sun, Jiacheng and Li, Zhenguo and Li, Chongxuan},
  booktitle = {International Conference on Learning Representations (ICLR)},
  volume    = {2025},
  pages     = {64972--65009},
  year      = {2025}
}

@inproceedings{nie2024smdm,
  title     = {Scaling up Masked Diffusion Models on Text},
  author    = {Nie, Shen and Zhu, Fengqi and Du, Chao and Pang, Tianyu and Liu, Qian and Zeng, Guangtao and Lin, Min and Li, Chongxuan},
  booktitle = {International Conference on Learning Representations (ICLR)},
  volume    = {2025},
  pages     = {82974--82997},
  year      = {2025}
}

@misc{bie2025llada2,
  title         = {{LLaDA2.0}: Scaling Up Diffusion Language Models to 100B},
  author        = {Bie, Tiwei and Cao, Maosong and Chen, Kun and Du, Lun and Gong, Mingliang and Gong, Zhuochen and Gu, Yanmei and Hu, Jiaqi and Huang, Zenan and Lan, Zhenzhong and Li, Chengxi and Li, Chongxuan and Li, Jianguo and Li, Zehuan and Liu, Huabin and Liu, Lin and Lu, Guoshan and Lu, Xiaocheng and Ma, Yuxin and Tan, Jianfeng and Wei, Lanning and Wen, Ji-Rong and Xing, Yipeng and Zhang, Xiaolu and Zhao, Junbo and Zheng, Da and Zhou, Jun and Zhou, Junlin and Zhou, Zhanchao and Zhu, Liwang and Zhuang, Yihong},
  year          = {2025},
  eprint        = {2512.15745},
  archivePrefix = {arXiv}
}

@misc{song2025seed,
  title         = {Seed Diffusion: A Large-Scale Diffusion Language Model with High-Speed Inference},
  author        = {Song, Yuxuan and Zhang, Zheng and Luo, Cheng and Gao, Pengyang and Xia, Fan and Luo, Hao and Li, Zheng and Yang, Yuehang and Yu, Hongli and Qu, Xingwei and Fu, Yuwei and Su, Jing and Zhang, Ge and Huang, Wenhao and Wang, Mingxuan and Yan, Lin and Jia, Xiaoying and Liu, Jingjing and Ma, Wei-Ying and Zhang, Ya-Qin and Wu, Yonghui and Zhou, Hao},
  year          = {2025},
  eprint        = {2508.02193},
  archivePrefix = {arXiv}
}

@misc{ye2025dreamvl,
  title         = {{Dream-VL} \& {Dream-VLA}: Open Vision-Language and Vision-Language-Action Models with Diffusion Language Model Backbone},
  author        = {Ye, Jiacheng and Gong, Shansan and Gao, Jiahui and Fan, Junming and Wu, Shuang and Bi, Wei and Bai, Haoli and Shang, Lifeng and Kong, Lingpeng},
  year          = {2025},
  eprint        = {2512.22615},
  archivePrefix = {arXiv}
}

@misc{swerdlow2025unified,
  title         = {Unified Multimodal Discrete Diffusion},
  author        = {Swerdlow, Alexander and Prabhudesai, Mihir and Gandhi, Siddharth and Pathak, Deepak and Fragkiadaki, Katerina},
  year          = {2025},
  eprint        = {2503.20853},
  archivePrefix = {arXiv}
}

@misc{zeng2025diffusionvl,
  title         = {{DiffusionVL}: Translating Any Autoregressive Models into Diffusion Vision Language Models},
  author        = {Zeng, Lunbin and Yao, Jingfeng and Liao, Bencheng and Tao, Hongyuan and Liu, Wenyu and Wang, Xinggang},
  year          = {2025},
  eprint        = {2512.15713},
  archivePrefix = {arXiv}
}

@inproceedings{jiang2025d2cache,
  title     = {{$d^2$Cache}: Accelerating Diffusion-Based {LLM}s via Dual Adaptive Caching},
  author    = {Jiang, Yuchu and Cai, Yue and Luo, Xiangzhong and Fu, Jiale and Wang, Jiarui and Liu, Chonghan and Yang, Xu},
  booktitle = {International Conference on Learning Representations (ICLR)},
  volume    = {2026},
  pages     = {137511--137531},
  year      = {2026}
}

@inproceedings{wu2025fastdllmv2,
  title     = {{Fast-dLLM v2}: Efficient Block-Diffusion {LLM}},
  author    = {Wu, Chengyue and Zhang, Hao and Xue, Shuchen and Diao, Shizhe and Fu, Yonggan and Liu, Zhijian and Molchanov, Pavlo and Luo, Ping and Han, Song and Xie, Enze},
  booktitle = {International Conference on Learning Representations (ICLR)},
  volume    = {2026},
  pages     = {128353--128370},
  year      = {2026}
}

@inproceedings{kang2025parallelbench,
  title     = {{ParallelBench}: Understanding the Trade-offs of Parallel Decoding in Diffusion {LLM}s},
  author    = {Kang, Wonjun and Galim, Kevin and Oh, Seunghyuk and Lee, Minjae and Zeng, Yuchen and Zhang, Shuibai and Hooper, Coleman and Hu, Yuezhou and Koo, Hyung and Cho, Nam Ik and Lee, Kangwook},
  booktitle = {International Conference on Learning Representations (ICLR)},
  volume    = {2026},
  pages     = {89900--89938},
  year      = {2026}
}

@inproceedings{wei2025slowfast,
  title     = {Accelerating Diffusion Large Language Models with {SlowFast} Sampling: The Three Golden Principles},
  author    = {Wei, Qingyan and Zhang, Yaojie and Liu, Zhiyuan and Zeng, Puyu and Wang, Yuxuan and Qi, Biqing and Liu, Dongrui and Zhang, Linfeng},
  booktitle = {International Conference on Learning Representations (ICLR)},
  volume    = {2026},
  pages     = {4340--4354},
  year      = {2026}
}

@inproceedings{wang2025d2f,
  title     = {Diffusion {LLM}s Can Do Faster-Than-{AR} Inference via Discrete Diffusion Forcing},
  author    = {Wang, Xu and Xu, Chenkai and Jin, Yijie and Jin, Jiachun and Zhang, Hao and Yu, Kai and Deng, Zhijie},
  booktitle = {International Conference on Learning Representations (ICLR)},
  volume    = {2026},
  pages     = {2282--2299},
  year      = {2026}
}

@misc{xiang2026cover,
  title         = {Stop the Flip-Flop: Context-Preserving Verification for Fast Revocable Diffusion Decoding},
  author        = {Xiang, Yanzheng and Wei, Lan and Yao, Yizhen and Zhu, Qinglin and Yan, Hanqi and Jin, Chen and Teare, Philip Alexander and Zhang, Dandan and Gui, Lin and Saseendran, Amrutha and He, Yulan},
  year          = {2026},
  eprint        = {2602.06161},
  archivePrefix = {arXiv}
}

@misc{huang2025maskprophet,
  title         = {Mask Tokens as Prophet: Fine-Grained Cache Eviction for Efficient {dLLM} Inference},
  author        = {Huang, Jianuo and Zhang, Yaojie and Yang, Yicun and Huang, Benhao and Qi, Biqing and Liu, Dongrui and Zhang, Linfeng},
  year          = {2025},
  eprint        = {2510.09309},
  archivePrefix = {arXiv}
}

@misc{zhang2026swordsman,
  title         = {Swordsman: Entropy-Driven Adaptive Block Partition for Efficient Diffusion Language Models},
  author        = {Zhang, Yu and Li, Xinchen and Zhou, Jialei and Ma, Hongnan and Wan, Zhongwei and Shi, Yiwei and Miao, Duoqian and Zhang, Qi and Cao, Longbing},
  year          = {2026},
  eprint        = {2602.04399},
  archivePrefix = {arXiv}
}

@misc{zhao2026anchors,
  title         = {Context Tokens are Anchors: Understanding the Repetition Curse in {dMLLM}s from an Information Flow Perspective},
  author        = {Zhao, Qiyan and Zhang, Xiaofeng and Chang, Shuochen and Chen, Qianyu and Yuan, Xiaosong and Chen, Xuhang and Liu, Luoqi and Zhang, Jiajun and Zhang, Xu-Yao and Wang, Da-Han},
  year          = {2026},
  eprint        = {2601.20520},
  archivePrefix = {arXiv}
}

@inproceedings{rohrbach2018chair,
  title     = {Object Hallucination in Image Captioning},
  author    = {Rohrbach, Anna and Hendricks, Lisa Anne and Burns, Kaylee and Darrell, Trevor and Saenko, Kate},
  booktitle = {Proceedings of the 2018 Conference on Empirical Methods in Natural Language Processing (EMNLP)},
  pages     = {4035--4045},
  doi       = {10.18653/v1/D18-1437},
  year      = {2018}
}

@inproceedings{liu2023llava,
  title     = {Visual Instruction Tuning},
  author    = {Liu, Haotian and Li, Chunyuan and Wu, Qingyang and Lee, Yong Jae},
  booktitle = {Advances in Neural Information Processing Systems (NeurIPS)},
  volume    = {36},
  pages     = {34892--34916},
  doi       = {10.52202/075280-1516},
  year      = {2023}
}

@inproceedings{yue2024mmmu,
  title     = {{MMMU}: A Massive Multi-discipline Multimodal Understanding and Reasoning Benchmark for Expert {AGI}},
  author    = {Yue, Xiang and Ni, Yuansheng and Zhang, Kai and Zheng, Tianyu and Liu, Ruoqi and Zhang, Ge and Stevens, Samuel and Jiang, Dongfu and Ren, Weiming and Sun, Yuxuan and Wei, Cong and Yu, Botao and Yuan, Ruibin and Sun, Renliang and Yin, Ming and Zheng, Boyuan and Yang, Zhenzhu and Liu, Yibo and Huang, Wenhao and Sun, Huan and Su, Yu and Chen, Wenhu},
  booktitle = {Proceedings of the {IEEE/CVF} Conference on Computer Vision and Pattern Recognition (CVPR)},
  pages     = {9556--9567},
  year      = {2024}
}

@inproceedings{liu2024mmbench,
  title     = {{MMBench}: Is Your Multi-modal Model an All-around Player?},
  author    = {Liu, Yuan and Duan, Haodong and Zhang, Yuanhan and Li, Bo and Zhang, Songyang and Zhao, Wangbo and Yuan, Yike and Wang, Jiaqi and He, Conghui and Liu, Ziwei and Chen, Kai and Lin, Dahua},
  booktitle = {European Conference on Computer Vision (ECCV)},
  volume    = {15064},
  pages     = {216--233},
  publisher = {Springer},
  doi       = {10.1007/978-3-031-72658-3_13},
  year      = {2024}
}

@inproceedings{chen2024m3cot,
  title     = {{$M^3$CoT}: A Novel Benchmark for Multi-Domain Multi-step Multi-modal Chain-of-Thought},
  author    = {Chen, Qiguang and Qin, Libo and Zhang, Jin and Chen, Zhi and Xu, Xiao and Che, Wanxiang},
  booktitle = {Proceedings of the 62nd Annual Meeting of the Association for Computational Linguistics (Volume 1: Long Papers)},
  pages     = {8199--8221},
  doi       = {10.18653/v1/2024.acl-long.446},
  year      = {2024}
}

@inproceedings{lee2024vlind,
  title     = {{VLind-Bench}: Measuring Language Priors in Large Vision-Language Models},
  author    = {Lee, Kang-il and Kim, Minbeom and Yoon, Seunghyun and Kim, Minsung and Lee, Dongryeol and Koh, Hyukhun and Jung, Kyomin},
  booktitle = {Findings of the Association for Computational Linguistics: NAACL 2025},
  pages     = {4129--4144},
  doi       = {10.18653/v1/2025.findings-naacl.231},
  year      = {2025}
}

\clearpage
\section*{Supplementary Material}
This supplement strengthens the empirical case for CD-APC along three axes: reproducibility, mechanism, and quality--efficiency trade-offs. Counterfactual image interventions show that visual gain tracks task-relevant token support; controlled attributions isolate candidate correction as the source of the M3CoT gain; and paired evaluations characterize how CCAW preserves quality while reducing computation.

\appendix

\section{Experimental Protocol and Reproducibility}

\subsection{Hardware and Software}

Experiments were run on two NVIDIA RTX PRO 6000 Blackwell Server Edition GPUs with 96~GB memory per GPU. The recorded driver version for the main sweep was 595.71.05. We used PyTorch 2.8.0 with CUDA 12.8, Transformers 4.46.0, and a fixed snapshot of VLMEvalKit (bundle \texttt{MMaDA\_DCD\_cloud\_bundle\_20260711}). Distributed inference used \texttt{torchrun --standalone --nproc-per-node=2}; data were sharded by rank and each GPU processed one sample at a time.

All MMaDA experiments used MMaDA-8B-MixCoT with the MagViT-v2 visual tokenizer at image resolution 512. Generation used \texttt{max\_new\_tokens=1024}, 512 denoising steps, and \texttt{block\_length=1024}. Semi-autoregressive decoding, KV caching, feature caching, and the auxiliary CV path were disabled. Decoding used deterministic argmax selection with temperature zero. Under the fixed software and hardware stack, independent MMBench reruns produced identical per-sample predictions and the same circular score (49.97).

MMBench prompts were constructed with the MMaDA chat template and included the hint and answer choices. LLaVABench used the original questions. MMBench was scored both by flat exact matching and by circular consistency over the pre-generated option permutations. LLaVABench was evaluated by the GPT-4o judge with four concurrent workers and up to three retries for failed requests; retries were used only to complete a judging run and were not counted as independent repetitions.

\begin{figure}[t]
\centering
\includegraphics[width=\linewidth]{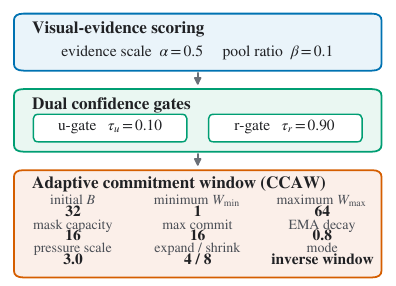}
\caption{Default CD-APC configuration, grouped by function. The sweep grid is listed separately in Section~\ref{sec:sensitivity}.}
\label{fig:hyperparameters}
\end{figure}

Wall-clock measurements use runner timestamps and include model loading. The observed MMBench throughput was approximately 2.3 seconds per sample per GPU. All method comparisons within a sweep used the same bundle and code path. We report paired sample-level uncertainty whenever predictions for both methods are available.

\subsection{Datasets and Statistical Procedures}

The counterfactual probe used 4,377 unique MMBench development samples and all 60 LLaVABench samples. The MMBench two-cycle file was generated before evaluation to avoid modifying answer permutations during inference. The file identifiers used for the experiments were \path{33af034601909d0a67496b8d8526e1e2} for MMBench and \path{d382a093f749a697820d3dadd61c8428} for LLaVABench.

For paired accuracy comparisons, confidence intervals and $p$ values were obtained by resampling complete examples, preserving the pairing between methods. Counterfactual image comparisons and LLaVABench regressions used 2,000 sample-level bootstrap replicates. MMBench logistic regressions used 500 sample-level bootstrap replicates. Repeated diffusion states from the same example were aggregated within that example before dataset-level inference, so long outputs or trajectories with more denoising states did not receive greater dataset-level weight.

\subsection{Hyperparameter Sensitivity}
\label{sec:sensitivity}

One-at-a-time sweeps used a separate grid around the default $\alpha=0.5,\beta=0.1$:
\[\begin{aligned}\alpha &: \{0.3,0.4,0.5,0.6,0.7\},\\ \beta &: \{0.05,0.1,0.2\}.\end{aligned}\]
Every point used the complete MMBench two-cycle evaluation and all 60 LLaVABench examples.

\begin{table}[!ht]
\centering
\footnotesize
\caption{MMaDA CD-APC sensitivity. Circ./flat gives the MMBench scores; $\Delta$ is the paired flat-score difference from the default with its 95\% CI and $p$ value.}
\label{tab:sensitivity}
\setlength{\tabcolsep}{1.8pt}
\begin{tabular}{@{}lclc@{}}
\toprule
Setting & Circ./flat & $\Delta$flat [95\% CI]; $p$ & LLaVA \\
\midrule
$\alpha=0.3$ & 50.06/57.38 & $+0.06$ [$-0.20,+0.31$]; .73 & 30.9 \\
$\alpha=0.4$ & 49.87/57.35 & $+0.02$ [$-0.14,+0.18$]; .89 & 31.8 \\
$\alpha=0.5$ & 49.97/57.33 & default & 32.1 \\
$\alpha=0.6$ & 49.76/57.27 & $-0.06$ [$-0.21,+0.10$]; .55 & 31.2 \\
$\alpha=0.7$ & 49.67/57.28 & $-0.05$ [$-0.25,+0.16$]; .75 & 31.5 \\
\midrule
$\beta=0.05$ & 49.87/57.21 & $-0.12$ [$-0.21,-0.03$]; .013 & 31.5 \\
$\beta=0.10$ & 49.97/57.33 & default & 32.1 \\
$\beta=0.20$ & 49.94/57.37 & $+0.05$ [$-0.07,+0.16$]; .54 & 31.9 \\
\bottomrule
\end{tabular}
\end{table}

Performance is stable across $\alpha\in[0.3,0.7]$ (maximum absolute flat-score difference: 0.06 points), placing $\alpha=0.5$ at a robust interior operating point. The default $\beta=0.1$ attains the strongest measured circular score and joint MMBench--LLaVABench result; $\beta=0.05$ lowers flat accuracy by 0.12 points, while $\beta=0.2$ remains close to the default. These sweeps support $(\alpha,\beta)=(0.5,0.1)$ as a stable default.

\section{Counterfactual Validation of Visual Gain}

\subsection{Image Interventions}

At a fixed decoding state $x_t$, an additional read-only probe evaluated four image conditions without changing the normal decoding state, commitment decisions, or CCAW state:
\begin{enumerate}
    \item the correct image $I$;
    \item the blocked branch used by CD-APC, which prevents direct output-to-image access;
    \item a mismatched image $I_{\mathrm{mis}}$ from another example in the same dataset; and
    \item a shuffled image $I_{\mathrm{shuf}}$, obtained by a deterministic permutation of the current image's VQ tokens.
\end{enumerate}
All branches shared the same prompt, partial decoded sequence, masked positions, candidate token, and valid text vocabulary. Mismatched images were selected by a deterministic derangement and checked by SHA256 identifier; shuffled images used a fixed per-example seed.

For the CD-APC candidate $y$ at state $x_t$, we measured
\begin{align}
g_{\mathrm{direct}} &= \log p(y\mid I,x_t)-\log p(y\mid \mathrm{blocked},x_t), \\
g_{\mathrm{mis}} &= \log p(y\mid I,x_t)-\log p(y\mid I_{\mathrm{mis}},x_t), \\
g_{\mathrm{shuf}} &= \log p(y\mid I,x_t)-\log p(y\mid I_{\mathrm{shuf}},x_t).
\end{align}
Probabilities were renormalized over the valid text vocabulary. Positive gain means that the correct image provides greater support for the evaluated token than the corresponding control condition.

We report three scopes. \emph{All-valid} contains candidate states whose positions lie within the final output length. Repeated states are first aggregated within an example. \emph{Commit} retains the state at which a position is actually committed. \emph{Alphanumeric commit} further excludes special tokens and retains decoded pieces containing at least one Unicode letter or digit; on MMBench, this scope mainly contains answer tokens.

\subsection{Correct Images Provide Greater Token Support}

\begin{figure*}[t]
\centering
\includegraphics[width=\textwidth]{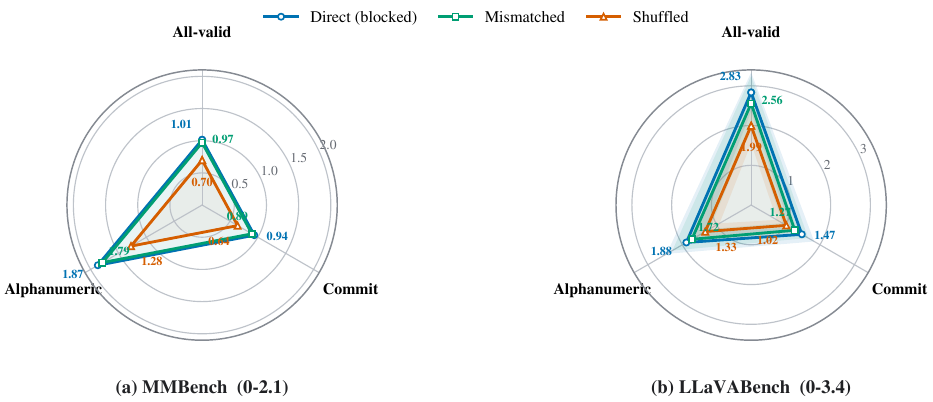}
\caption{Counterfactual visual gain by dataset and scope. Points show means and shaded bands show 95\% sample-bootstrap confidence intervals. Dataset-specific zero-based radial scales expand within-dataset contrasts (MMBench: 0--2.1; LLaVABench: 0--3.4); polygon areas are therefore not comparable across panels. Every interval lies above zero.}
\label{fig:counterfactual_gain}
\end{figure*}

Figure~\ref{fig:counterfactual_gain} shows that the correct image provides greater support than all three controls. On MMBench all-valid states, the proportions of tokens with positive gain were 82.53\%, 58.64\%, and 70.00\% for blocked, mismatched, and shuffled controls, respectively. Gains were particularly large for alphanumeric commits, showing that image sensitivity is concentrated at positions carrying answer content rather than being uniformly distributed across the trajectory.

The blocked comparison is the internal contrast used by CD-APC, whereas mismatched and shuffled gains were not used to choose the CD-APC candidate. Their consistent positive values therefore provide a counterfactual check that is not entirely explained by the internal blocked-branch selection rule.

\subsection{Visual Gain Predicts MMBench Correctness Beyond Confidence}

MMBench correctness was defined by exact match between the model prediction and the answer. For each gain, we fitted
\begin{equation}
\operatorname{logit} P(\mathrm{correct})
=\beta_0+\beta_c z(c_{\mathrm{base}})+\beta_g z(g),
\end{equation}
where $z(\cdot)$ denotes sample-level standardization. Table~\ref{tab:gain-correctness} reports the multiplicative change in the odds of a correct answer for a one-standard-deviation increase in gain.

\begin{table}[!ht]
\centering
\footnotesize
\caption{Incremental prediction of MMBench correctness after controlling for base confidence.}
\label{tab:gain-correctness}
\begin{tabular}{lcc}
\toprule
Gain & Odds ratio [95\% CI] & Marginal effect \\
\midrule
$g_{\mathrm{direct}}$ & 1.331 [1.237, 1.429] & +6.45 pp \\
$g_{\mathrm{mis}}$ & 1.230 [1.162, 1.309] & +4.71 pp \\
$g_{\mathrm{shuf}}$ & 1.117 [1.055, 1.191] & +2.53 pp \\
\bottomrule
\end{tabular}
\end{table}

The corresponding commit-scope odds ratios were 1.375, 1.267, and 1.130. Thus, visual gain is not a duplicate of base confidence: it provides incremental information about closed-set correctness, with the strongest effect for the direct blocked-branch contrast used by CD-APC.

On LLaVABench, visual gain captures a complementary property: after controlling for confidence, the all-valid coefficient for $g_{\mathrm{direct}}$ was $-0.795$ with a 95\% confidence interval of $[-1.112,-0.424]$. Larger gains identify visually demanding open-ended prompts that receive less support from language priors. Together, the two datasets position $g$ as a state-level image-dependence signal: it predicts correctness in the closed-ended setting and identifies visual demand in open-ended generation.

\section{Controlled Attribution of Commitment Timing and Content}

\subsection{Attribution Configurations}

The controlled attribution suite isolates three effects under a shared decoding implementation:
\begin{itemize}
    \item \textbf{$g=0$:} the visual gain is set to zero while the gate structure and other decoding controls are retained;
    \item \textbf{Timing-only:} real gain controls commitment eligibility and timing, but the committed token is forced to the visual branch's base top-1 candidate; and
    \item \textbf{Full CAFW:} real gain may affect both commitment timing and the committed candidate under the fixed-window CAFW decoder.
\end{itemize}
The Original decoder provides the common reference. This design separates visual timing from contrastive candidate correction without removing the gate thresholds themselves.

\subsection{M3CoT Accuracy Attribution}

\begin{table}[!ht]
\centering
\footnotesize
\caption{Paired M3CoT accuracy comparisons. Variant accuracies were 35.98 (Original), 36.54 ($g=0$), 36.93 (Timing-only), and 38.74 (Full CAFW).}
\label{tab:gate-attribution}
\setlength{\tabcolsep}{3.5pt}
\begin{tabular}{@{}lcc@{}}
\toprule
Comparison & Difference [95\% CI] & $p$ \\
\midrule
$g=0$ vs. Original & +0.56 [$-0.86,+1.98$] & 0.47 \\
Timing-only vs. Original & +0.95 [$-0.47,+2.37$] & 0.21 \\
Timing-only vs. $g=0$ & +0.39 [$-0.47,+1.25$] & 0.45 \\
Full CAFW vs. $g=0$ & \textbf{+2.20 [$+0.82,+3.62$]} & \textbf{0.003} \\
Full CAFW vs. Timing-only & \textbf{+1.81 [$+0.60,+3.02$]} & \textbf{0.004} \\
\bottomrule
\end{tabular}
\end{table}

All variants used the same decoder skeleton and paired tests. On M3CoT, Full CAFW improves over $g=0$ and Timing-only by 2.20 and 1.81 points, respectively, whereas timing alone remains close to both Original and $g=0$. MMBench flat scores (56.64, 56.76, and 57.10) and LLaVABench scores ($32.0\pm0.3$ versus $31.7\pm0.2$) remain stable across the controlled variants. The M3CoT gain therefore concentrates on candidate correction: a 0.18\% flip rate is sufficient to produce a 1.81-point advantage through downstream propagation.

\subsection{Trajectory and Efficiency Analysis}

The trajectory comparison covers approximately 1.187 million aligned position-state pairs. Table~\ref{tab:trajectory} summarizes timing and content differences between the controlled variants.

\begin{table}[!ht]
\centering
\footnotesize
\caption{Trajectory statistics for real gain relative to $g=0$.}
\label{tab:trajectory}
\begin{tabular}{lr}
\toprule
Metric & Result \\
\midrule
Mean commit-time shift & $-1.56$ steps \\
Earlier / later / unchanged & 15.8 / 6.9 / 77.3\% \\
First-snapshot commit Jaccard & 0.832 \\
NFE, Timing-only / $g=0$ & 47.8 / 50.4 \\
Candidate flip rate & 0.18\% \\
Candidate flip rate at commits & 0.04\% \\
Final-token disagreement & 2.72\% \\
\bottomrule
\end{tabular}
\end{table}

Real gain advances commitment by 1.56 steps on average and reduces forward evaluations by 5.2\% in Timing-only. The effect is selective and bidirectional: consistent positions tend to commit earlier and divergent cases later; 22.7\% change time. Sparse candidate changes together with larger final disagreement support a propagation mechanism in which selected corrections influence later positions.

\section{Additional Paired Performance Analyses}

\subsection{MMaDA on M3CoT}

\begin{table}[!ht]
\centering
\footnotesize
\caption{Paired MMaDA/M3CoT comparisons on 2,318 examples.}
\label{tab:m3cot-main}
\setlength{\tabcolsep}{1.7pt}
\begin{tabular}{@{}lccc@{}}
\toprule
Comparison & Difference & 95\% CI & $p$ \\
\midrule
Fixed vs. Original & +2.76 & $[+1.47,+4.10]$ & 0.0004 \\
CD-APC vs. Original & +2.63 & $[+1.25,+3.97]$ & 0.0001 \\
CD-APC vs. Fixed & $-0.13$ & $[-0.43,+0.13]$ & 0.55 \\
CD-APC vs. w/o r & +4.53 & $[+3.15,+5.95]$ & $<0.0001$ \\
CD-APC vs. w/o u & 0.00 & -- & 1.0 \\
\bottomrule
\end{tabular}
\end{table}

The r-gate is decisive on M3CoT: removing it reduces accuracy from 38.61 to 34.08. The u-gate is inactive in this high-confidence regime, where mean base confidence is approximately 0.995, far above $\tau_u=0.1$. This controlled comparison attributes the observed M3CoT gain to the relevance criterion.

\subsection{MMaDA on MMBench}

\begin{table}[!ht]
\centering
\footnotesize
\caption{Paired MMBench comparisons under circular and flat scoring. Differences are in percentage points.}
\label{tab:mmbench-main}
\setlength{\tabcolsep}{2pt}
\begin{tabular}{@{}lcc@{}}
\toprule
Comparison & $\Delta$ [95\% CI] & $p$ \\
\midrule
\multicolumn{3}{l}{\textbf{Circular scoring}} \\
CD-APC vs. Original & +0.80 [$+0.07,+1.53$] & .039 \\
CD-APC vs. DCD & +0.34 [$-0.34,+1.03$] & .36 \\
CD-APC vs. Fixed & +0.34 [$-0.07,+0.78$] & .13 \\
Fixed vs. Original & +0.46 [$-0.27,+1.17$] & .24 \\
DCD vs. Original & +0.46 [$-0.14,+1.05$] & .16 \\
\midrule
\multicolumn{3}{l}{\textbf{Flat scoring}} \\
CD-APC vs. Original & +1.21 [$+0.70,+1.72$] & $<.0001$ \\
CD-APC vs. DCD & +0.64 [$+0.16,+1.11$] & .011 \\
CD-APC vs. Fixed & +0.32 [$+0.03,+0.61$] & .039 \\
Fixed vs. Original & +0.89 [$+0.37,+1.39$] & .0007 \\
DCD vs. Original & +0.58 [$+0.16,+1.01$] & .009 \\
\bottomrule
\end{tabular}
\end{table}

CD-APC improves over Original under both scoring protocols. Flat scoring also resolves a +0.32-point CD-APC advantage over the fixed-window variant; the corresponding circular-score difference is +0.34 points.

\subsection{Open-Ended Generation}

On MMaDA/LLaVABench, CD-APC and the fixed-window visual method improve over Original by 19.1 $[13.8,25.0]$ and 18.4 $[12.9,24.3]$ points. Their intervals overlap, with CD-APC attaining the highest observed score among the two visual-control variants (32.1 versus 31.3) and matching Official DCD (32.2).

On LaViDA/LLaVA-COCO, both visual-control variants transfer strongly: the fixed-window method improves over Original from 54.55 to 80.34 (+25.79 $[19.01,32.58]$, $p<0.0001$), while CD-APC reaches 72.15 (+17.60 $[11.77,23.69]$, $p<0.0001$). The fixed window is stronger by 8.19 points in this open-ended LaViDA setting, separating cross-backbone transfer of visual control from the backbone-specific quality--efficiency behavior of CCAW.

\section{Quality--Efficiency Characterization of CCAW}

\begin{table}[!ht]
\centering
\scriptsize
\caption{Controlled CCAW versus fixed-$W{=}16$. Quality and NFE are reported as CCAW/fixed; tests are paired.}
\label{tab:ccaw-controlled}
\setlength{\tabcolsep}{2pt}
\begin{tabular}{@{}lccc@{}}
\toprule
Dataset / gain & Quality C/F & $\Delta$; test & NFE C/F \\
\midrule
LLaVABench / $g=0$ & 32.30/32.11 & $+0.19$; $p=.75$ & 117.8/205.2 \\
MMBench flat / $g=0$ & 56.73/56.64 & $+0.09$; $p=.60$ & 8.4/8.7 \\
M3CoT / visual $g$ & 38.61/38.74 & $-0.13$ [$-.43,+.13$] & 40.8/42.1 \\
\bottomrule
\end{tabular}
\end{table}

With $g=0$, CCAW cuts LLaVABench NFE by 42.6\% while preserving the paired quality score; MMBench shows the same quality-preserving pattern with a smaller NFE reduction. Output length is stable at 110.8 versus 111.9 words. Window percentiles $(p_{10},p_{50},p_{90})$ are $(16,17,53)$ on LLaVABench and $(20,36,48)$ on MMBench, indicating operation near the lower bound with intermittent expansion. The efficiency gain therefore arises from adaptive scheduling itself.

\begin{table}[!ht]
\centering
\scriptsize
\caption{Fixed-window sweep with visual gain enabled and the adaptive CCAW reference. Entries give quality / two-GPU wall time in minutes. Missing full-run times are estimated by scaling the main-paper per-sample wall times to the complete evaluation.}
\label{tab:fixed-window-pareto}
\setlength{\tabcolsep}{3pt}
\begin{tabular}{@{}lcc@{}}
\toprule
Setting & LLaVA rel. / min & MMB circ. / min \\
\midrule
Original & 12.9 / 12.1 & 49.17 / 766 \\
Fixed $W=16$ & 31.3 / 27.5 & 49.62 / 169 \\
Fixed $W=32$ & 30.7 / 12.5 & 49.60 / 91 \\
Fixed $W=64$ & 28.2 / 10.8 & 49.58 / 54 \\
CCAW ($16\!\rightarrow\!64$) & \textbf{32.1} / 12 & \textbf{49.97} / 165 \\
\bottomrule
\end{tabular}
\end{table}

Larger static windows reduce time but monotonically lower observed quality. CCAW has the highest observed quality; on MMBench it approaches fixed-$W{=}16$ runtime, while $W=64$ is about three times faster but 0.39 points lower. On LLaVABench, CCAW improves over $W=16$ by 0.8 points while cutting wall time by more than half, and exceeds $W=64$ by 3.9 points at similar cost. Together with Table~\ref{tab:ccaw-controlled}, this establishes CCAW as the strongest quality--efficiency operating point among the tested schedules.

\section{Judge Repeatability}

The same LLaVABench predictions were submitted to three independent GPT-4o judging runs. Table~\ref{tab:judge-repeat} reports the resulting relative scores.

\begin{table}[!ht]
\centering
\footnotesize
\caption{Independent GPT-4o judging repetitions on fixed predictions.}
\label{tab:judge-repeat}
\begin{tabular}{lcccc}
\toprule
Variant & Run 0 & Run 1 & Run 2 & Mean $\pm$ SD \\
\midrule
$g=0$ & 32.1 & 32.2 & 31.7 & $32.00\pm0.26$ \\
Timing-only & 31.9 & 31.8 & 31.5 & $31.73\pm0.21$ \\
\bottomrule
\end{tabular}
\end{table}

Across three runs, Timing-only minus $g=0$ averaged $-0.27$, with paired bootstrap interval $[-2.47,+2.03]$. The observed difference lies within judge variation and leaves the attribution result unchanged.

\clearpage
\onecolumn
\section{Qualitative Case Studies}

\begin{figure}[!ht]
\centering
\includegraphics[width=0.90\textwidth]{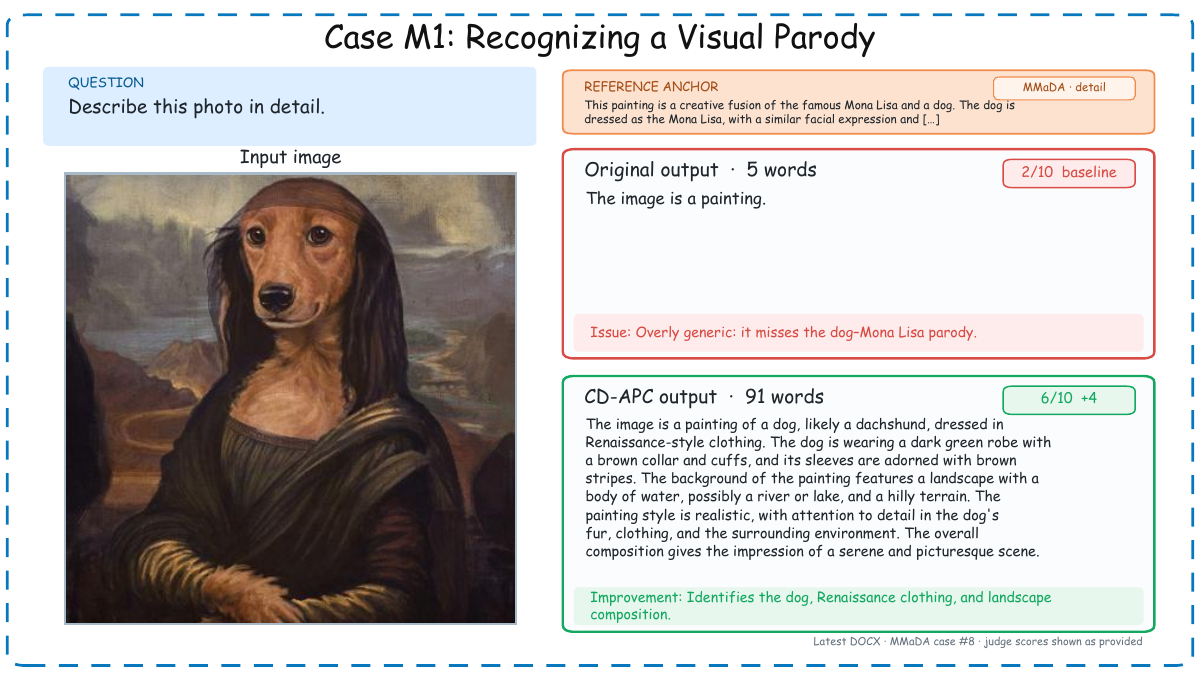}\\[2pt]
\includegraphics[width=0.90\textwidth]{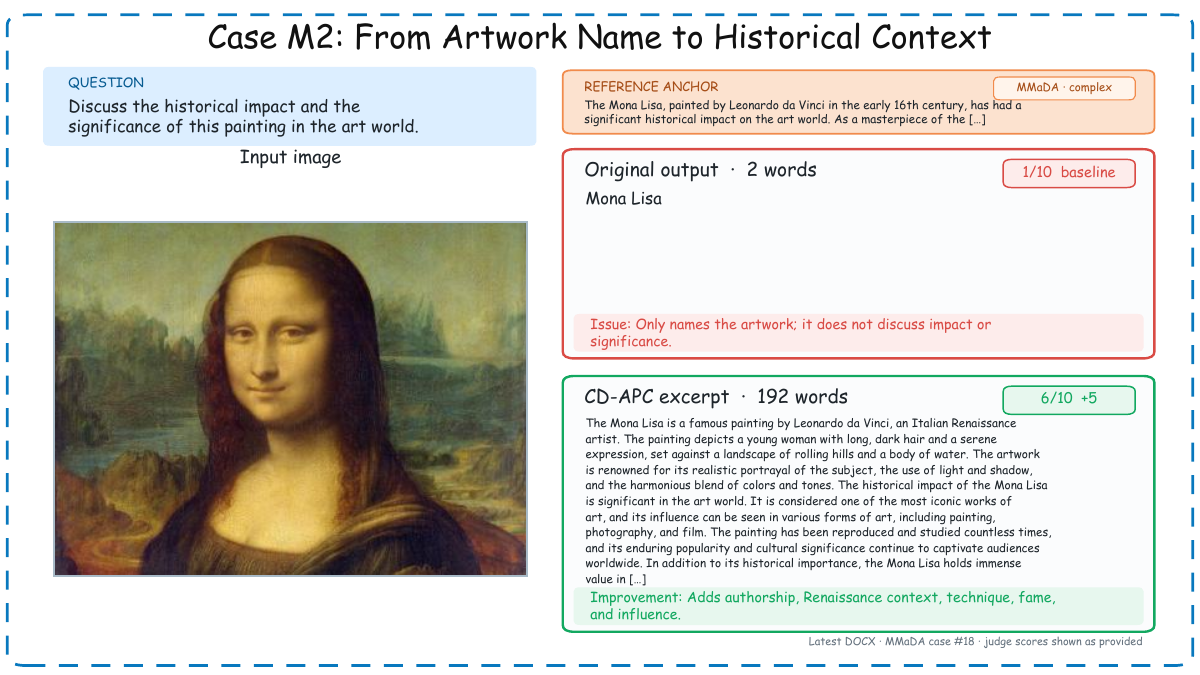}
\caption{MMaDA cases M1--M2: visual-parody recognition and historical-context generation.}
\label{fig:cases-m12}
\end{figure}

\begin{figure}[!ht]
\centering
\includegraphics[width=0.90\textwidth]{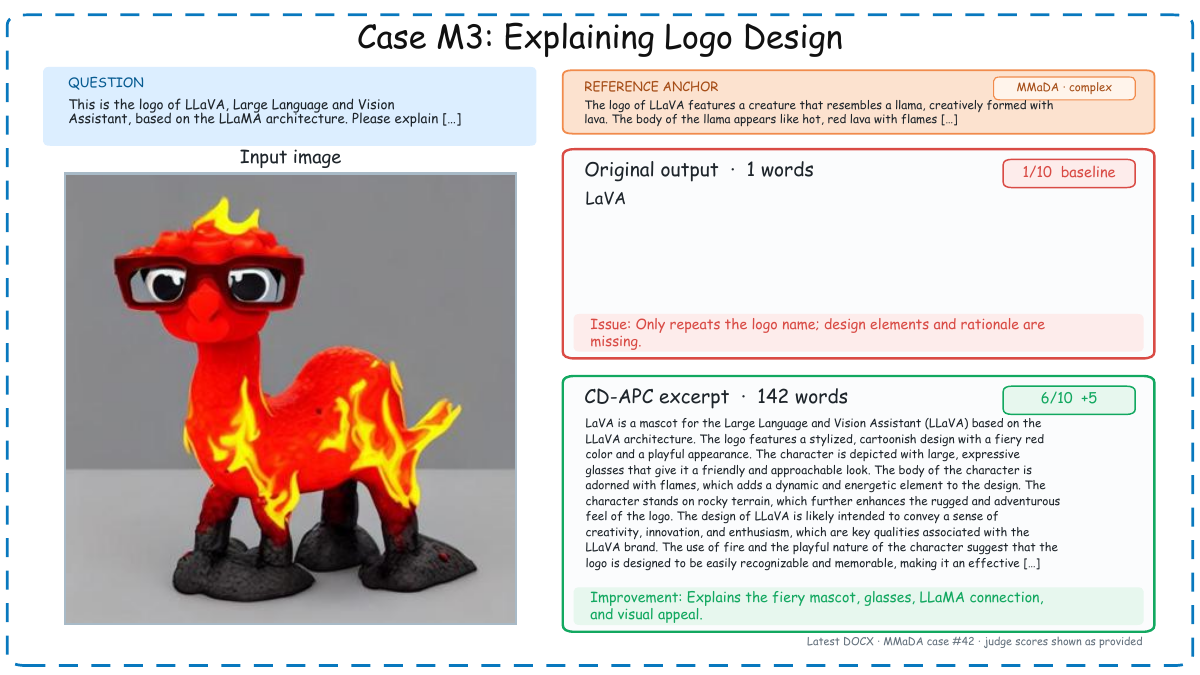}\\[2pt]
\includegraphics[width=0.90\textwidth]{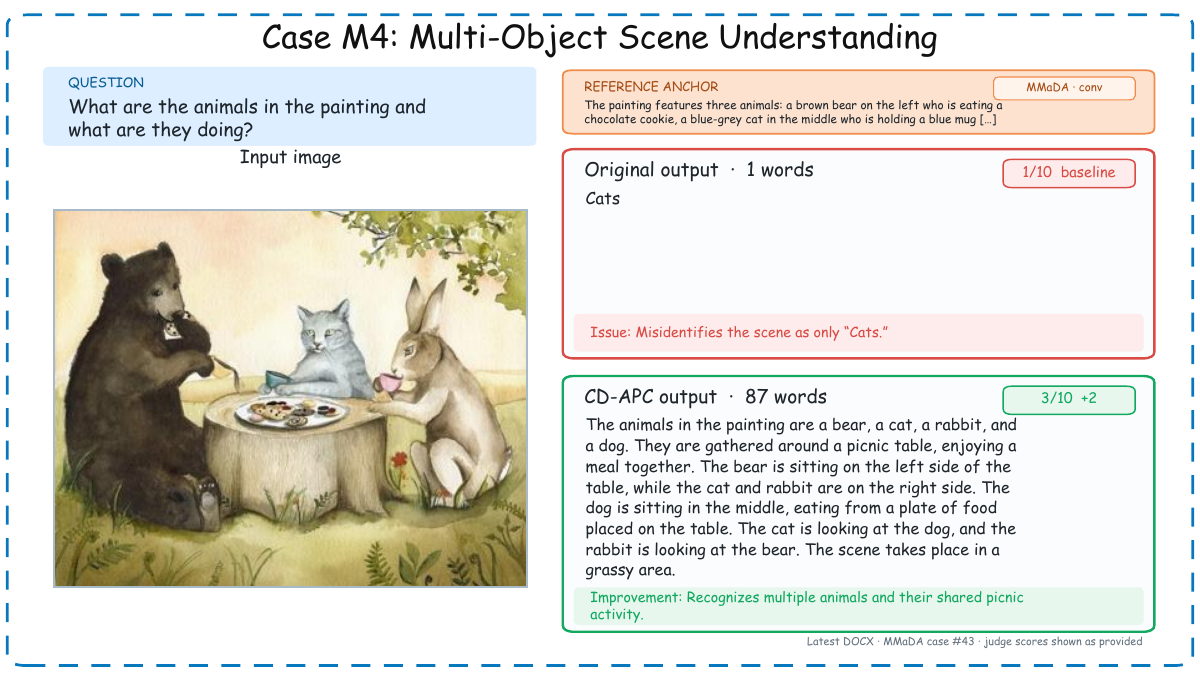}
\caption{MMaDA cases M3--M4: logo explanation and multi-object scene understanding.}
\label{fig:cases-m34}
\end{figure}

\begin{figure}[!ht]
\centering
\includegraphics[width=0.90\textwidth]{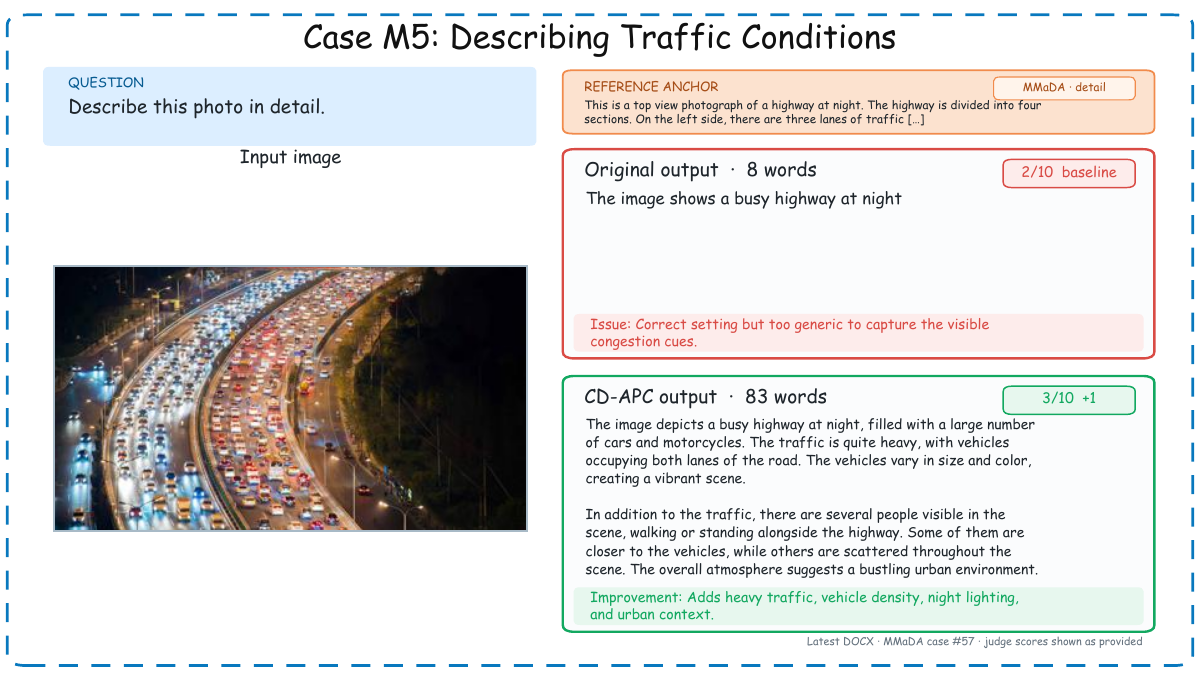}\\[2pt]
\includegraphics[width=0.90\textwidth]{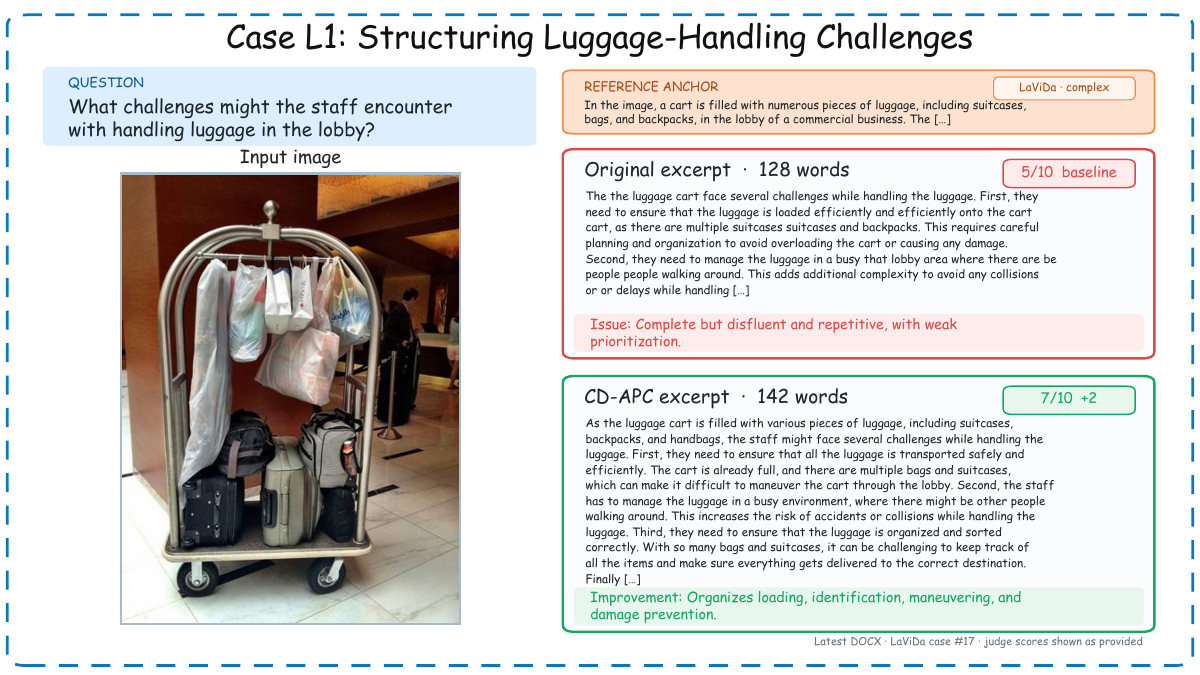}
\caption{Cases M5 and L1: detailed traffic description and structured luggage-handling analysis.}
\label{fig:cases-m5l1}
\end{figure}

\begin{figure}[!ht]
\centering
\includegraphics[width=0.90\textwidth]{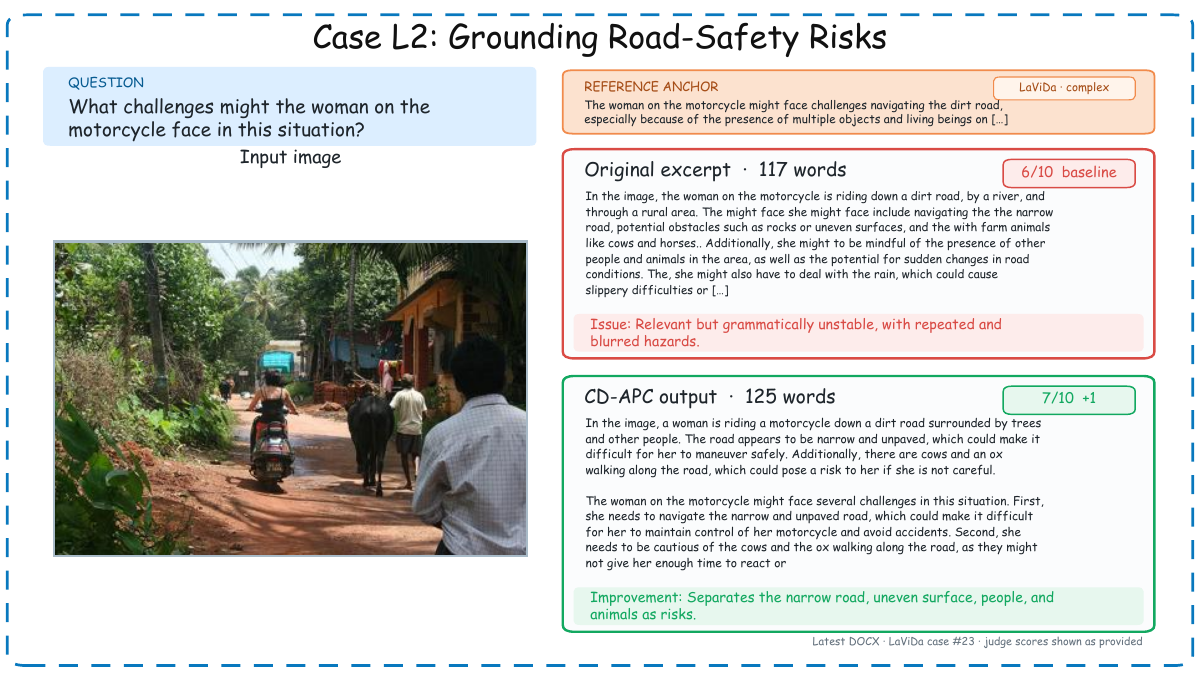}\\[2pt]
\includegraphics[width=0.90\textwidth]{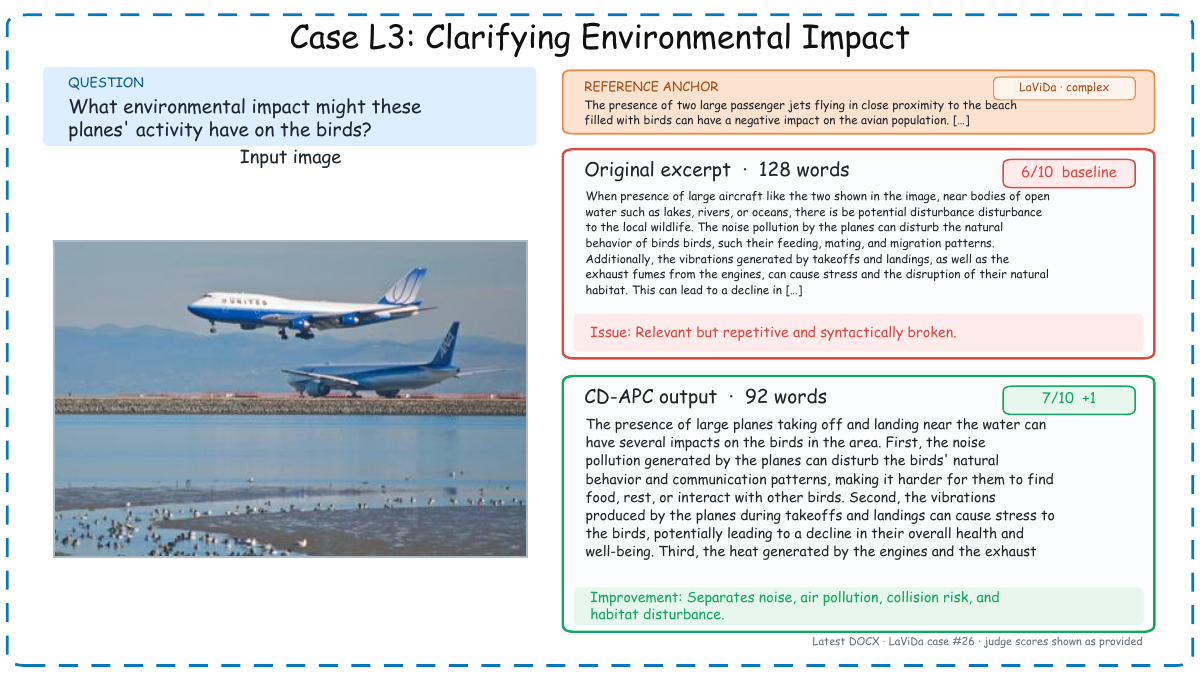}
\caption{LaViDa cases L2--L3: grounded road-safety risks and environmental-impact reasoning.}
\label{fig:cases-l23}
\end{figure}

\begin{figure}[!ht]
\centering
\includegraphics[width=0.90\textwidth]{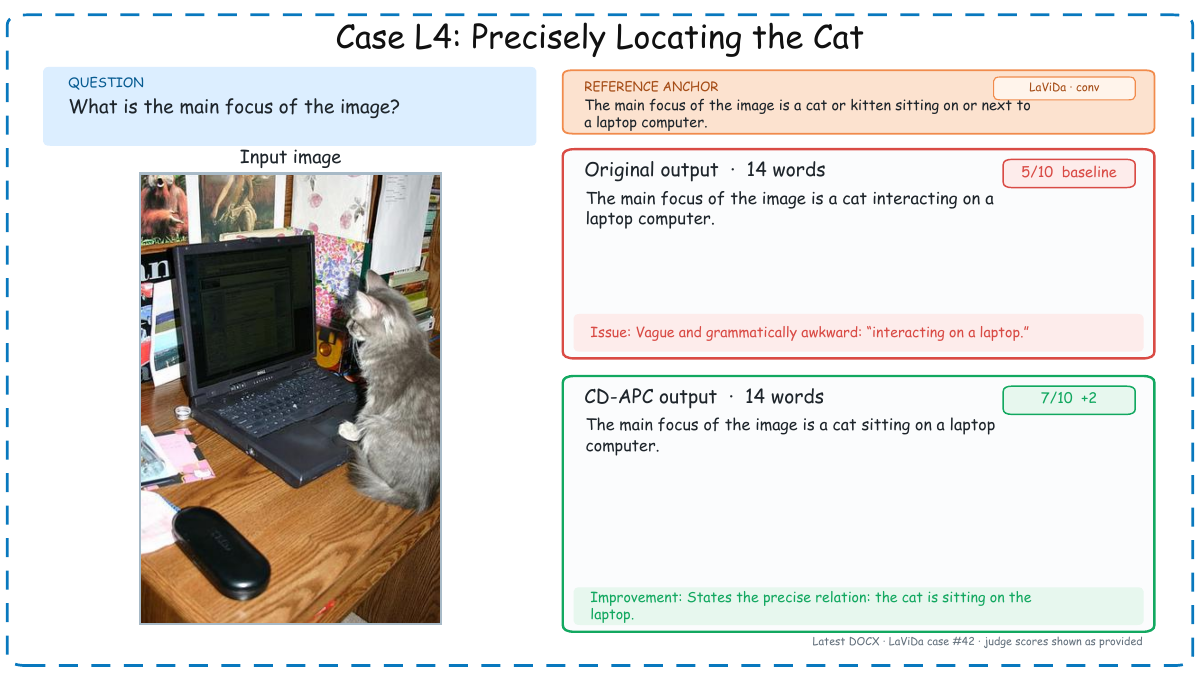}\\[2pt]
\includegraphics[width=0.90\textwidth]{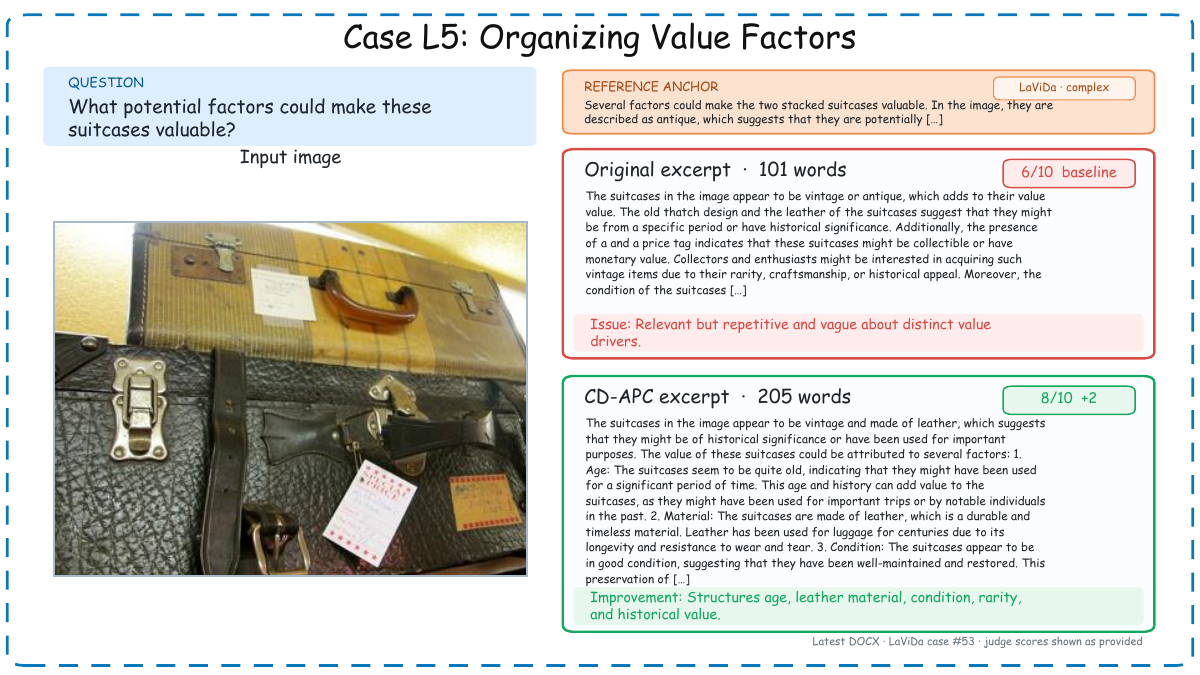}
\caption{LaViDa cases L4--L5: precise spatial grounding and organized value-factor analysis.}
\label{fig:cases-l45}
\end{figure}

Figures~\ref{fig:cases-m12}--\ref{fig:cases-l45} present ten representative LLaVABench cases across MMaDA and LaViDa. Each case pairs the input and reference anchor with the Original and CD-APC outputs, making the recovery from short, generic, or disfluent responses directly visible. The examples complement the aggregate results by showing improvements in visual specificity, multi-object grounding, response structure, and image-conditioned reasoning.

\clearpage

\clearpage

\clearpage

\clearpage

\end{document}